\documentclass[conference]{IEEEtran}
\IEEEoverridecommandlockouts
\usepackage{siunitx}
\usepackage{graphicx,xfp}
\usepackage{multicol}
\usepackage{multirow}
\usepackage{color}
\usepackage{xcolor}
\usepackage{booktabs}
\usepackage{graphicx}
\usepackage{nicefrac}
\usepackage{amsmath}
\usepackage{array}
\usepackage{hyperref}

\usepackage{algorithm}
\usepackage{algpseudocode}

\usepackage{cite}
\usepackage{amsmath,amssymb,amsfonts}
\usepackage{graphicx}
\usepackage{textcomp}

\usepackage{amssymb}

\usepackage{amsmath,amsfonts}
\usepackage{array}
\usepackage{textcomp}
\usepackage{stfloats}
\usepackage{url}
\usepackage{hyperref}
\usepackage{verbatim}
\usepackage{graphicx}
\usepackage{cite}
\usepackage{xcolor}
\usepackage{multirow}
\usepackage{graphicx,xfp}
\usepackage{booktabs,threeparttable}
\usepackage{tikz}
\usepackage{tikzscale}
\usepackage{comment}
\usepackage{caption}
\usepackage{subcaption}
\usepackage{pgfplots}
\usepgfplotslibrary{groupplots}
\usetikzlibrary{positioning}
\usetikzlibrary{external}
\usepackage{cite}
\usepackage{amsmath,amssymb,amsfonts}
\usepackage{graphicx}
\usepackage{textcomp}
\usepackage{xcolor}
\def\BibTeX{{\rm B\kern-.05em{\sc i\kern-.025em b}\kern-.08em
    T\kern-.1667em\lower.7ex\hbox{E}\kern-.125emX}}

\begin{document}

\title{Exploring The Potential Of GANs In Biological Sequence Analysis}

\author{\IEEEauthorblockN{Taslim Murad, Sarwan Ali and Murray Patterson}
\IEEEauthorblockA{\textit{Department of Computer Science, Georgia State University},
Atlanta, USA \\
\{tmurad2,sali85\}@student.gsu.edu, mpatterson30@gsu.edu}
}

\maketitle

\begin{abstract}
Biological sequence analysis is an essential step toward building a deeper understanding of the underlying functions, structures, and behaviors of the sequences. It can help in identifying the characteristics of the associated organisms, like viruses, etc., and building prevention mechanisms to eradicate their spread and impact, as viruses are known to cause epidemics which can become pandemics globally. 
New tools for biological sequence analysis are provided by machine learning (ML) technologies to effectively analyze the functions and structures of the sequences. However, these ML-based methods undergo challenges with data imbalance, generally associated with biological sequence datasets, which hinders their performance. Although various strategies are present to address this issue, like the SMOTE algorithm, which create synthetic data, however, they focus on local information rather than the overall class distribution. In this work, we explore a novel approach to handle the data imbalance issue based on Generative Adversarial Networks (GANs) which use the overall data distribution. GANs are utilized to generate synthetic 
data that closely resembles the real one, thus this generated data can be employed to enhance the ML models' performance by eradicating the class imbalance problem for biological sequence analysis. We perform 3 distinct classification tasks by using 3 different sequence datasets (Influenza A Virus, PALMdb, VDjDB) and our results illustrate that GANs can improve the overall classification performance. 

\end{abstract}


\section{Introduction}\label{sec:intro}

Biological sequences usually refer to nucleotides or amino acids-based sequences and their analysis can provide detailed information about the functional and structural behaviors of the corresponding viruses which are usually responsible for causing diseases, for example, Flu~\cite{das2012antivirals}, Covid-19~\cite{pedersen2020sars}, etc. This information is very useful in building prevention mechanisms, like drugs~\cite{rognan2007chemogenomic}, vaccines~\cite{dong2007sequence}, etc., and to control the disease spread, eliminate the negative impacts, and do virus spread surveillance.  

Influenza A Virus (IAV) is such an example, which is responsible for causing a highly contagious respiratory illness that can significantly threaten global public health. 
As the Centers for Disease Control and Prevention Center (CDC)~\footnote{\url{https://www.cdc.gov/flu/weekly/index.htm}} reports that so far this season, there have been at least 25 million illnesses, 280,000 hospitalizations, and 17,000 deaths from Flu in the United States. Therefore, identifying and tracking the evolution of IAV accurately is a vital step in the fight against this virus. Classification of IAV is an essential task in this aspect as it can provide valuable information on the origin, evolution, and spread of the virus.
Moreover, the identification of the viral taxonomy can further enrich its understanding, like the viral polymerase palmprint sequence of a virus is utilized to determine its taxonomy (species generally)~\cite{babaian2022ribovirus}. A polymerase palmprint is a unique sequence of amino acids located at the thumb subunit of the viral RNA-dependent polymerase. Furthermore, examining the antigen specificities based on the T-cell receptor sequences can provide beneficial information regarding solving numerous problems of both basic and applied immunology research.

Many traditional sequence analysis methods follow phylogeny-based techniques ~\cite{hadfield2018a,minh_2020_iqtree2} to identify sequence homology and predict disease transmission. However, the availability of large-size sequence data exceeds the computational limit of such techniques.

Moreover, the application of ML approaches for performing biological sequence analysis is a popular research topic these days ~\cite{chen2022istrf,yang2022identification}. The ability of ML methods to determine the sequence's biological functions makes them desirable to be employed for sequence analysis. Additionally, ML models can also determine the relationship between the primary structure of the sequence and its biological functions. Like ~\cite{chen2022istrf} built a Random Forest-based algorithm to classify sucrose transporter (SUT) protein, ~\cite{yang2022identification} designed a novel tool for Protein-protein interactions data and functional analysis, ~\cite{zhang2023pseu} developed a new ML model to identify RNA pseudo-uridine modification sites. 
ML-based biological sequence analysis approaches can be categorized into feature-engineering-based methods~\cite{kuzmin2020machine,ma2020phylogenetic}, kernel-based methods~\cite{ghandi2014enhanced}, neural network-based techniques~\cite{shen2018wasserstein,xie2016unsupervised}, and pre-trained deep learning models~\cite{heinzinger2019modeling,strodthoff2020udsmprot}.
However, extrinsic factors limit the performance of ML-based techniques and one such major factor is data imbalance, as in the case of biological sequences the data is generally imbalanced because the number of negative samples is much larger than that of positive samples ~\cite{zhang2019balance}. ML models can obtain the best results when the dataset is balanced while unbalanced data will greatly affect the training of machine learning models and their application in real-world scenarios ~\cite{abd2013review}.

In this paper, we explore the idea to improve the performance of ML methods for biological sequence analysis by eradicating the data imbalance challenge using Generative Adversarial Networks (GANs). Our method leverages the strengths of GANs to effectively analyze these sequences, with the potential to have significant implications for virus surveillance and tracking, as well as the development of new antiviral strategies.

Our contributions to this work are as follows:
\begin{enumerate}
    \item We explore the idea 
    to do biological sequence classification using Generative Adversarial Networks (GANs).
    \item We show that usage of GANs improves predictive performance by eliminating the data imbalance challenge. 
    \item We demonstrated the potential implications of the proposed approach for virus surveillance and tracking, and for the development of new antiviral strategies. 
\end{enumerate}

The rest of the paper is organized as follows: Section~\ref{sec_related_work} contains the related work. The proposed approach details are enlisted in Section~\ref{sec_proposed_approach}. The datasets used in the experiments along with the ML models and evaluation metrics information is provided in Section~\ref{sec_experimental_setup}. Section~\ref{sec_results_discussion} highlights the experimental results and their discussion. Finally, the paper is concluded in Section~\ref{sec_conclusion}.

\section{Related Work}\label{sec_related_work}

The combination of biological sequence analysis and ML models has gained quite a lot of attention among researchers in recent years~\cite{chen2022istrf,yang2022identification}. As a biological sequence consists of a long string of characters corresponding to either nucleotides or amino acids, it needs to be transformed into a numerical form to make it compatible with the ML model. Various numerical embedding generation mechanisms are proposed to extract features from the biological sequences~\cite{kuzmin2020machine,strodthoff2020udsmprot,shen2018wasserstein}.

Some of the popular embedding generation techniques use the underlying concept of $k$-mer to compute the embeddings. Like ~\cite{ali2021spike2vec} uses the $k$-mers frequencies to get the vectors, ~\cite{ma2020phylogenetic,ali2022pwm2vec} combine position distribution information and $k$-mers frequencies for getting the embeddings. Other approaches~\cite{shen2018wasserstein,xie2016unsupervised} are employing neural networks to obtain the feature vectors. Moreover, kernel-based methods~\cite{ghandi2014enhanced} and pre-trained deep learning models-based methods~\cite{heinzinger2019modeling,strodthoff2020udsmprot} also play a vital role in generating the embeddings. Although all these techniques illustrate promising analysis results, however, they haven't mentioned anything about dealing with data imbalance issues, which if handled properly will yield performance improvement.

Furthermore, another set of methods tackles the class imbalance challenge with the aim to enhance overall analytical performance. They use resampling techniques at the data level by either oversampling the minority class or undersampling the majority class. For instance, ~\cite{chen2022istrf} uses Borderline-SMOTE algorithm~\cite{han2005borderline}, an oversampling approach, to balance the feature set of the sucrose transporter (SUT) protein dataset. However, as Borderline-SMOTE is based on the k nearest neighbor algorithm so it has high time complexity, is susceptible to noise data, and cannot make good use of the information of the majority samples~\cite{xiaolong2019over}. Similarly, ~\cite{zhao2008protein} does protein classification by handling the data imbalance using a hybrid sampling algorithm that combines both ensemble classifier and over-sampling techniques, KernelADASYN~\cite{tang2015kerneladasyn} employs a kernel-based adaptive synthetic over-sampling approach to deal with data imbalance. However, these methods don't utilize the overall data distribution, rather only are based on local information~\cite{douzas2018effective}. 

\section{Proposed Approach}\label{sec_proposed_approach}
In this section, we will discuss our idea of exploring GANs to obtain analytical performance improvement for biological sequences in detail. As our input sequence data consists of string sequences representing amino acids, they need to be transformed into numerical representations in order to operate GANs on them. For that purpose, we use $4$ distinct and effective numerical feature generation methods which are described below.

\subsection{Spike2Vec~\cite{ali2021spike2vec}}
Spike2Vec generates the feature embedding by computing the $k$-mers of a sequence. As $k$-mers are known to preserve the ordering information of the sequence. $K$-mers represent a set of consecutive substrings of length $k$ driven from a sequence. For s sequence with length $N$, the total number of its $k$-mers will be $N - k + 1$. This method devises the feature vector for a sequence by capturing the frequencies of its $k$-mers. To further deal with the curse of dimensionality issue, Spike2Vec uses random Fourier features (RFF) to map data to a randomized low-dimensional feature space. We use $k=3$ to get the embeddings.

\subsection{PWM2Vec~\cite{ali2022pwm2vec}}
\begin{figure}[h!]
  \centering
  \includegraphics[scale=0.30]{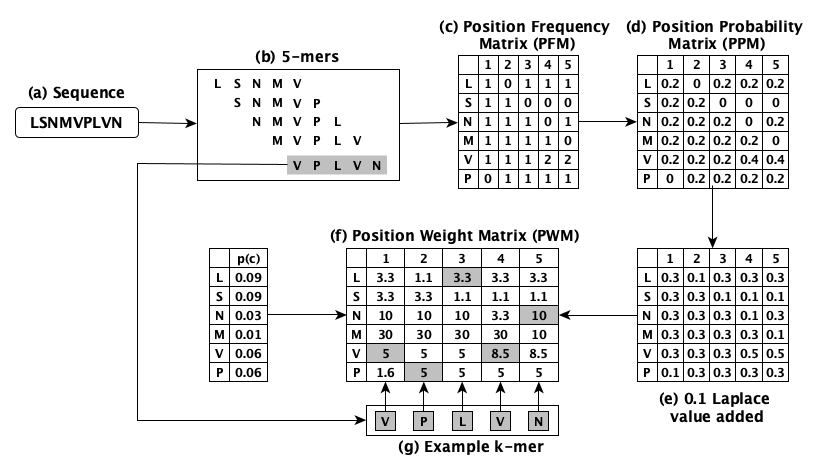}
 \caption{The workflow of PWM2Vec method for a given sequence.}
  \label{fig_pwm2vec_workflow}
\end{figure}

This method works by using the concept of $k$-mers to get the numerical form of the biological sequences, however, rather than utilizing constant frequency values of the $k$-mers, it assigns weights to each amino acid of the $k$-mers and employs these weights to generate the embeddings. The position weight matrix (PWM) is used to determine the weights. PWM2Vec considers the relative importance of amino acids along with preserving the ordering information. The workflow of this method is illustrated in Figure~\ref{fig_pwm2vec_workflow}. Our experiments use $k=3$ to obtain the embeddings. 

\subsection{Minimizer}
\begin{figure}[h!]
  \centering
  \includegraphics[scale=0.37]{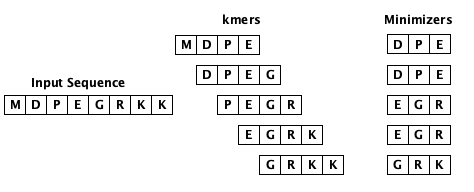}
 \caption{The workflow of getting minimizers from an input sequence. }
  \label{fig_minimizer_workflow}
\end{figure}

This approach is based on the utility of minimizers~\cite{roberts-2004-minimizer} ($m$-mer) to get the feature vectors of sequences. The minimizer is extracted from a $k$-mer and it is a $m$ length lexicographically smallest (in both forward and backward order) substring of consecutive alphabets from the $k$-mer. Note that $m<k$. The workflow of computing minimizers for a given input sequence is shown in Figure~\ref{fig_minimizer_workflow}. This approach intends to eliminate the redundancy issue associated with $k$-mers, hence improving the storage and computation cost. Our experiments used $k=9$ and $m=3$ to generate the embeddings.

After getting the numerical embeddings of the biological sequences using the methods mentioned above, we further utilize these embeddings to train our GAN model. This model has two parts, a generator model and a discriminator model. Each discriminator and generator model consists of two inner dense layers with ReLu activation functions (each followed by a batch-normalization layer) and a final dense layer. In the discriminator, the final dense layer has a Sigmoid activation function while the generator has a Softmax activation function. The generator's output has dimensions the same as the input data, as it synthesis the data, while the discriminator yields a binary scalar value to indicate if the generated data is fake or real. 

\begin{figure}[h!]
  \centering
  \includegraphics[scale=0.43]{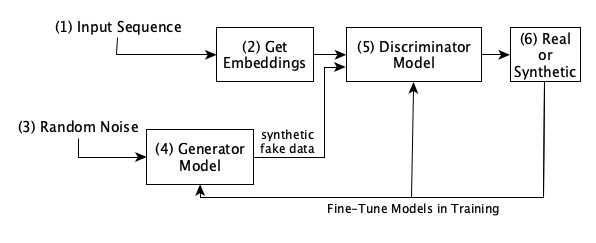}
 \caption{The workflow of the GAN model. }
  \label{fig_GANS_workflow}
\end{figure}

The GAN model is trained using cross-entropy loss function, ADAM optimizer, $32$ batch size, and $1000$ iterations. The steps followed to obtain the synthetic data after the training GAN model is illustrated in the Algorithm~\ref{algo_gans}. As given in the algorithm, firstly generator and discriminator models are created in steps 1-2. Then the discriminator model is complied for training with cross-entropy loss and ADAM optimizer in step 3. After that, the count and length of synthetic sequences along with the number of training epochs and batch size are mentioned in steps 4-6. Then the training of the models is happening in 7-12 steps, where each of the models is fine-tuned for the given number of iterations. Once the GAN model is trained, its generator part is employed to synthesize new embedding data which resembles real-world data. This synthesized data can eliminate the data imbalance problem, improving the analytical performance. Moreover, the workflow of GAN is shown in Figure~\ref{fig_GANS_workflow}.

\begin{algorithm}[h!]
	\caption{GAN Model.}
        \label{algo_gans}
	\begin{algorithmic}[1]
	\Statex \textbf{Input:} Set of Sequences $S$, $ganCnt$
	\Statex \textbf{Output:} GANs based sequences $S'$
        \State $m\_gen \gets generator()$ \Comment{generator model}

        \State $m\_dis \gets discriminator()$ \Comment{discriminator model}
        \State $m\_dis.compile(loss=CE, opt=ADAM)$

        \State $seqLen \gets len(S[0])$ \Comment{len of each $S'$ sequence}
        \State $iter \gets 1000$
        \State $batch\_size \gets 32$

        \For{$i$ in $iter$}
            \State $noise \gets \Call{random}{ganCnt,seqLen}$
            \State $S' \gets m\_gen.predict(noise)$  \Comment{get GAN sequences}

            \State $m\_dis.backward(m\_dis.loss)$ \Comment{fine-tune $m\_dis$}
            \State $m\_gen.backward(m\_gen.loss)$ \Comment{fine-tune $m\_gen$}

        \EndFor
        
        \State return($S'$)

	\end{algorithmic}
\end{algorithm}

\section{Experimental Setup}\label{sec_experimental_setup}
This section highlights the details of the datasets used to conduct the experiments along with the information about the classification models and their respective evaluation metrics to report the performance. 
All experiments are carried out on Intel (R) Core i5 system with a 2.40 GHz processor and $32$ GB memory.

\subsection{Dataset Statistics}
We use $3$ different datasets to evaluate our suggested method. 
A detailed description of each of the dataset is given as follow,

\subsubsection{Influenza A Virus}
We are using the influenza A virus sequence dataset belonging to two kinds of subtypes "H1N1'' and "H3N2'' extracted from ~\cite{BV-BRC} website. This data contains $222,450$ sequences in total with $119,100$ sequences belonging to the H1N1 subtype and $103,350$ to the H2N3 subtype. The detailed statics for this dataset is shown in Table~\ref{tab_data_statistics}. We use these two subtypes as labels to classify the Influenza A virus in our experiments. 

\begin{table}[h!]
    \centering
    \caption{Dataset Statistics of each dataset used in our experiments.}
    \resizebox{0.49\textwidth}{!}{
    \begin{tabular}{@{\extracolsep{4pt}}cccp{2cm}ccc}
    \toprule
    & & & & \multicolumn{3}{c}{Sequence Length} \\
    \cmidrule{5-7}
    Name & $\vert$ Sequences $\vert$ & Classes & Goal & Min & Max & Average \\
    \midrule \midrule
        \multirow{2}{*}{Influenza A Virus}  & \multirow{2}{*}{222450}  & \multirow{2}{*}{2} & Virus Subtypes Classification & \multirow{2}{*}{11} & \multirow{2}{*}{71} & \multirow{2}{*}{68.60} \\
        \midrule
        \multirow{2}{*}{PALMdb} & \multirow{2}{*}{124908} & \multirow{2}{*}{18} & Virus Species Classification & \multirow{2}{*}{53} & \multirow{2}{*}{150} & \multirow{2}{*}{130.83} \\
        \midrule
        \multirow{2}{*}{VDjDB} & \multirow{2}{*}{78344} & \multirow{2}{*}{17} & Antigen Species Classification & \multirow{2}{*}{7} & \multirow{2}{*}{20} & \multirow{2}{*}{12.66} \\
         \bottomrule
    \end{tabular}
    }
    \label{tab_data_statistics}
\end{table}

\subsubsection{PALMdb}
The PALMdb~\cite{babaian2022ribovirus,edgar2022petabase} dataset consists of viral polymerase palmprint sequences which can be classified species-wise. This dataset is created by mining the public sequence databases using the palmscan algorithm. It has $124,908$ sequences corresponding to $18$ different virus species. The distribution of these species is given in Table~\ref{tab:palmdb_count} and more detailed statistics are shown in Table~\ref{tab_data_statistics}. We use the species name as a label to do the classification of the PALMdb sequences. 

\begin{table}[h!]
    \centering
    \caption{Species-wise distribution of PALMdb dataset.}
    \resizebox{0.49\textwidth}{!}{
    \begin{tabular}{p{4cm}l|ll}
    \toprule
         Species Name & Count & Species Name & Count \\
    \midrule \midrule
        Avian orthoavulavirus 1 & 2353 & Chikungunya virus & 2319 \\
        Dengue virus & 1627 & Hepacivirus C & 29448 \\
        Human orthopneumovirus & 3398 & Influenza A virus & 47362 \\
        Influenza B virus & 8171 & Lassa mammarenavirus & 1435 \\
         Middle East respiratory syndrome-related coronavirus &  1415 & Porcine epidemic diarrhea virus & 1411 \\
         Porcine reproductive and respiratory syndrome virus & 2777 & Potato virus Y & 1287 \\
        Rabies lyssavirus & 4252 & Rotavirus A & 4214 \\
        Turnip mosaic virus & 1109 & West Nile virus & 5452 \\
        Zaire ebolavirus & 4821 & Zika virus & 2057 \\
    \bottomrule
    \end{tabular}
    }
    \label{tab:palmdb_count}
\end{table}

\subsubsection{VDjDB}
VDJdb is a curated dataset of T-cell receptor sequences with known antigen specificities ~\cite{bagaev2020vdjdb}. This dataset
consists of 58,795 human TCRs and 3,353 mouse TCRs. More than half of the examples are TRBs (n=
36,462) with the remainder being TRAs (n= 25,686). It has total $78,344$ sequences belonging to $17$ unique antigen species. The distribution of the sequence among the antigen species is shown in Table~\ref{tab:vdjdb_count} and further details of the dataset are given in Table~\ref{tab_data_statistics}. We use this data to perform the antigen species classification. 

Our code and preprocessed datasets are available online for reproducibility~\footnote{Available in the published version}.

\begin{table}[h!]
    \centering
    \caption{Antigen Species-wise distribution of VDjDB dataset.}
    \begin{tabular}{ll|ll}
    \toprule
         Antigen Species Name & Count & Antigen Species Name & Count \\
    \midrule \midrule
        CMV & 37357 & DENV1 & 180 \\
        DENV3/4 & 177 & EBV & 11026 \\
        HCV & 840 & HIV-1 &  3231 \\
        HSV-2 & 154 & HTLV-1 & 232 \\
        HomoSapiens & 4646 & InfluenzaA & 14863 \\
        LCMV & 141 & MCMV & 1463 \\
        PlasmodiumBerghei & 243 & RSV & 125 \\
        SARS-CoV-2 & 758 & SIV & 2119 \\
        YFV & 789 & & \\
    \bottomrule
    \end{tabular}
    \label{tab:vdjdb_count}
\end{table}

\subsection{Data Visualization}
\begin{figure}[h!]
  \centering
  \begin{subfigure}{.15\textwidth}
    \centering
    \includegraphics[scale=0.06]{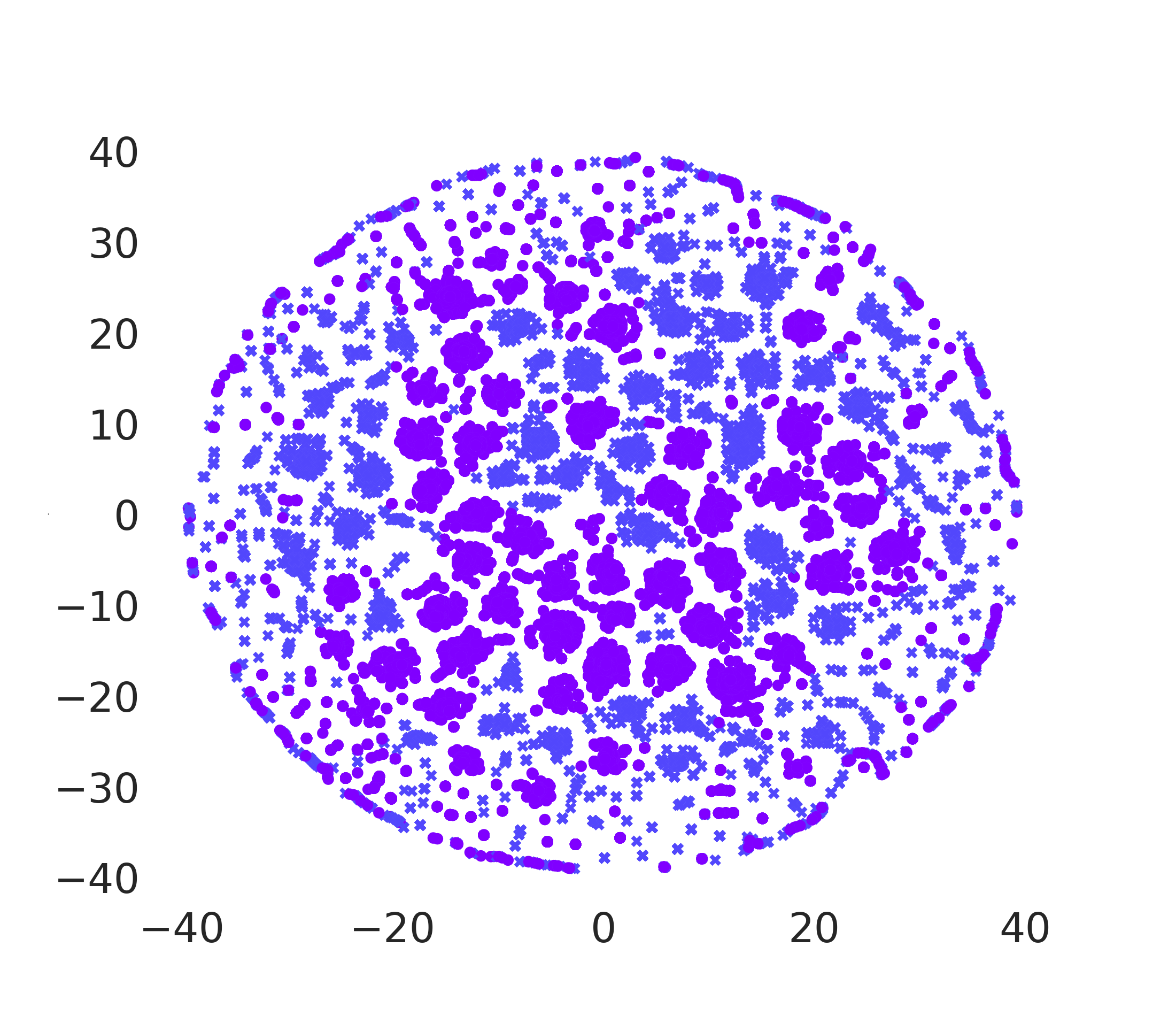}
    \caption{Spike2Vec}
    \label{fig_tsne_flu_spike2vec}
  \end{subfigure}%
  \begin{subfigure}{.15\textwidth}
    \centering
    \includegraphics[scale=0.06]{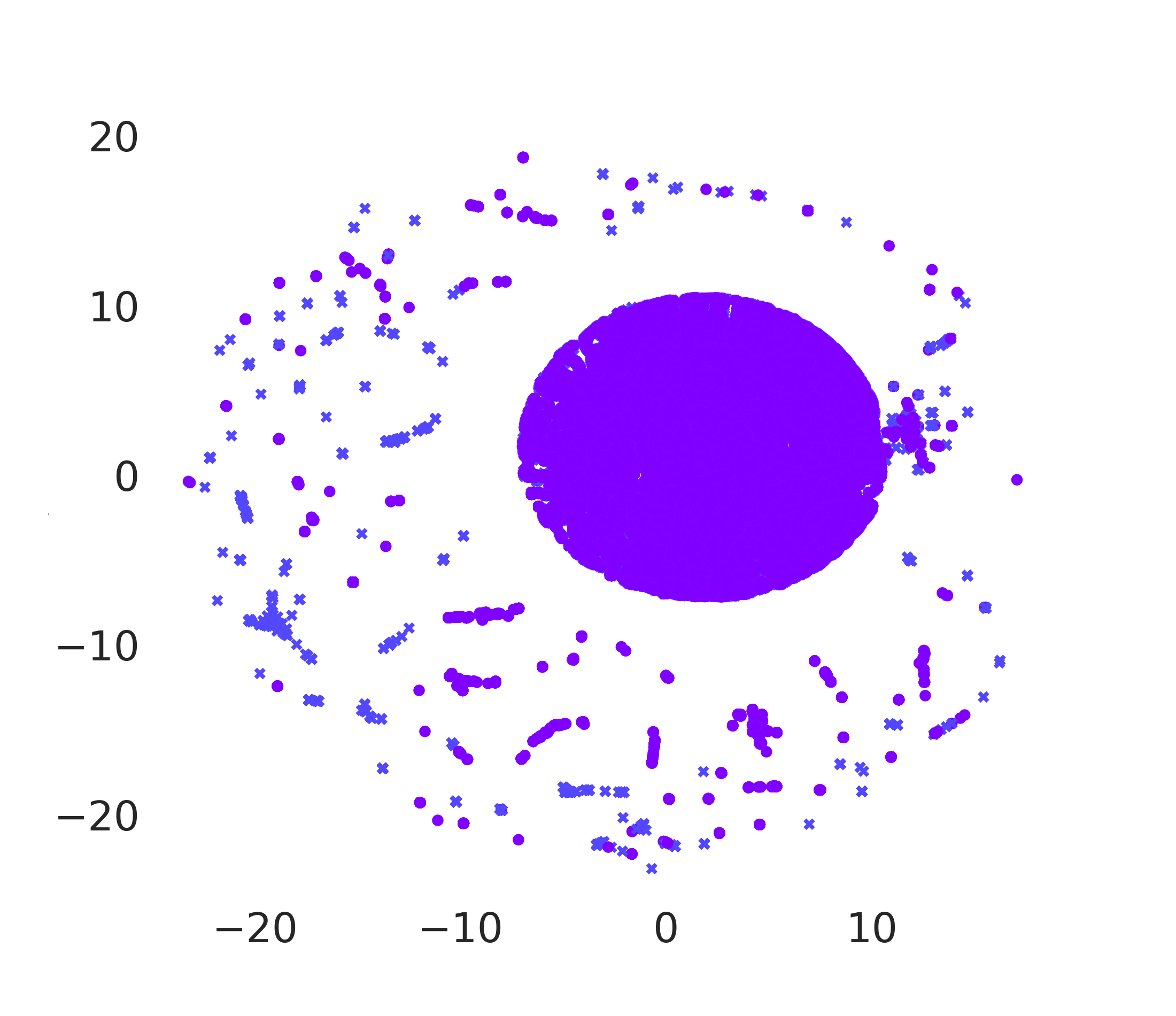}
    \caption{PWM2Vec}
    \label{fig_tsne_flu_pwm2vec}
  \end{subfigure}%
  \begin{subfigure}{.15\textwidth}
    \centering
    \includegraphics[scale=0.06]{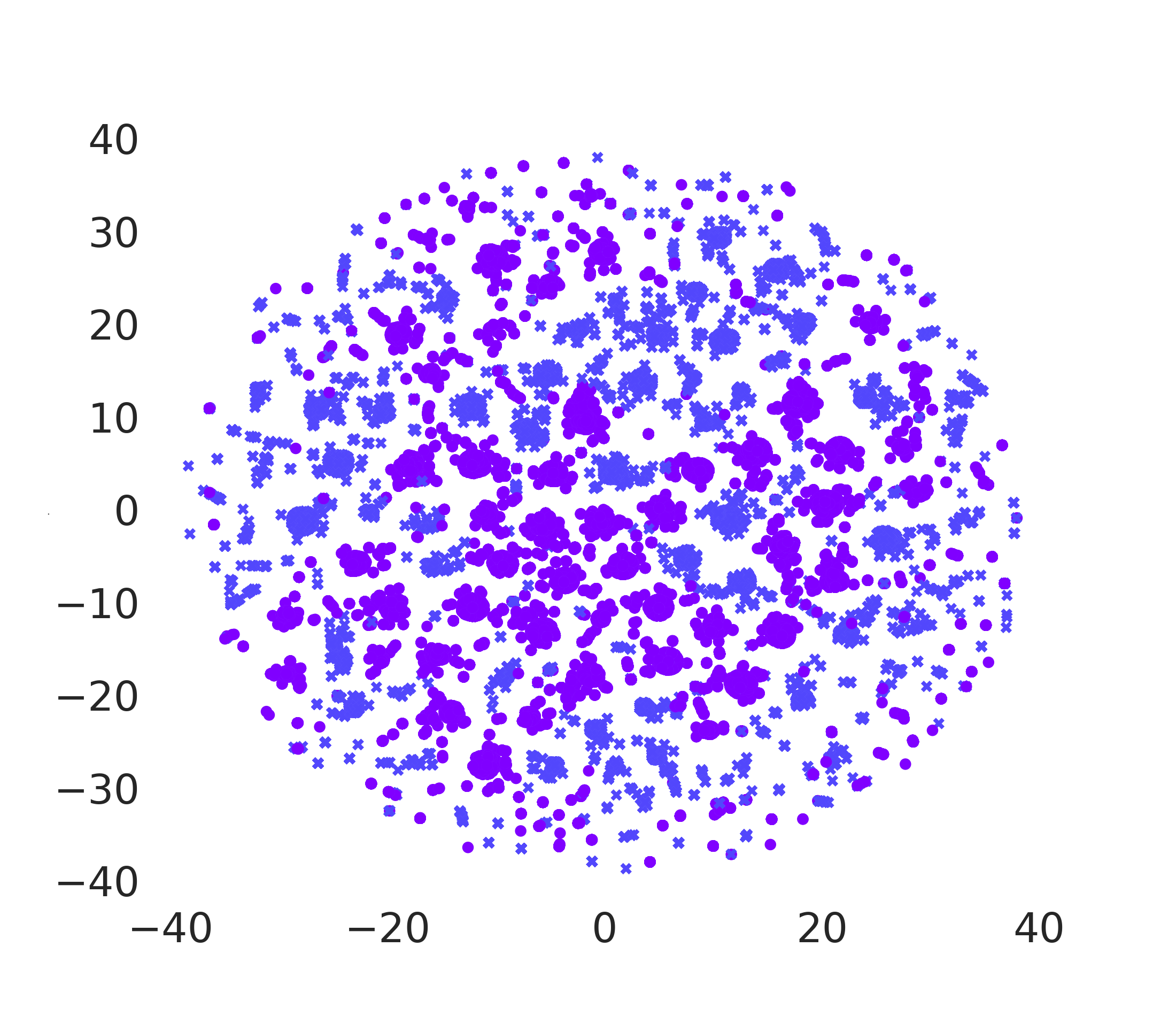}
    \caption{Minimizer}
    \label{fig_tsne_flu_minimizer}
  \end{subfigure}
  \\
  \begin{subfigure}{.15\textwidth}
    \centering
    \includegraphics[scale=0.06]{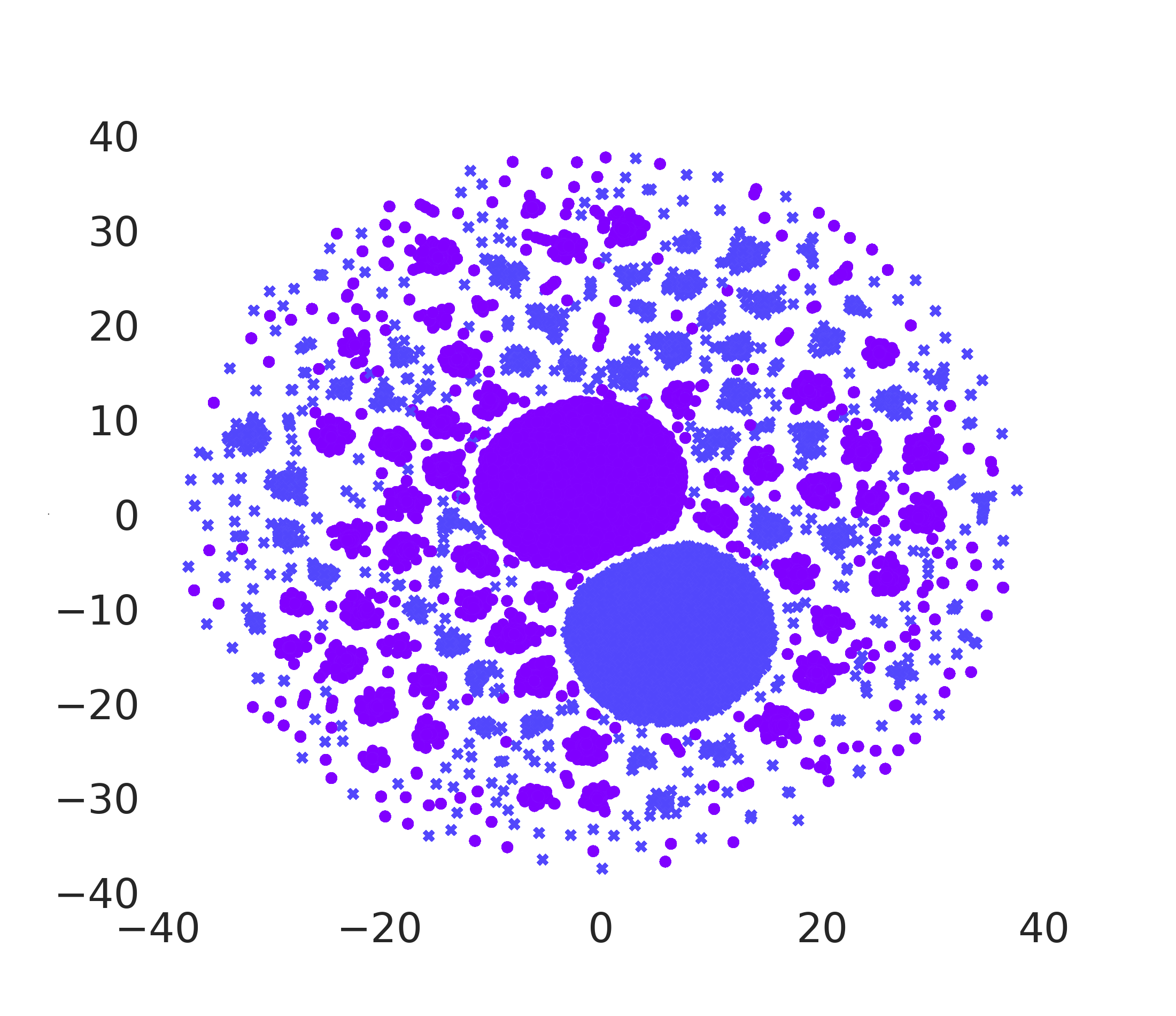}
    \caption{G-Spike2Vec}
    \label{fig_tsne_flu_gspike2vec}
  \end{subfigure}%
  \begin{subfigure}{.15\textwidth}
    \centering
    \includegraphics[scale=0.06]{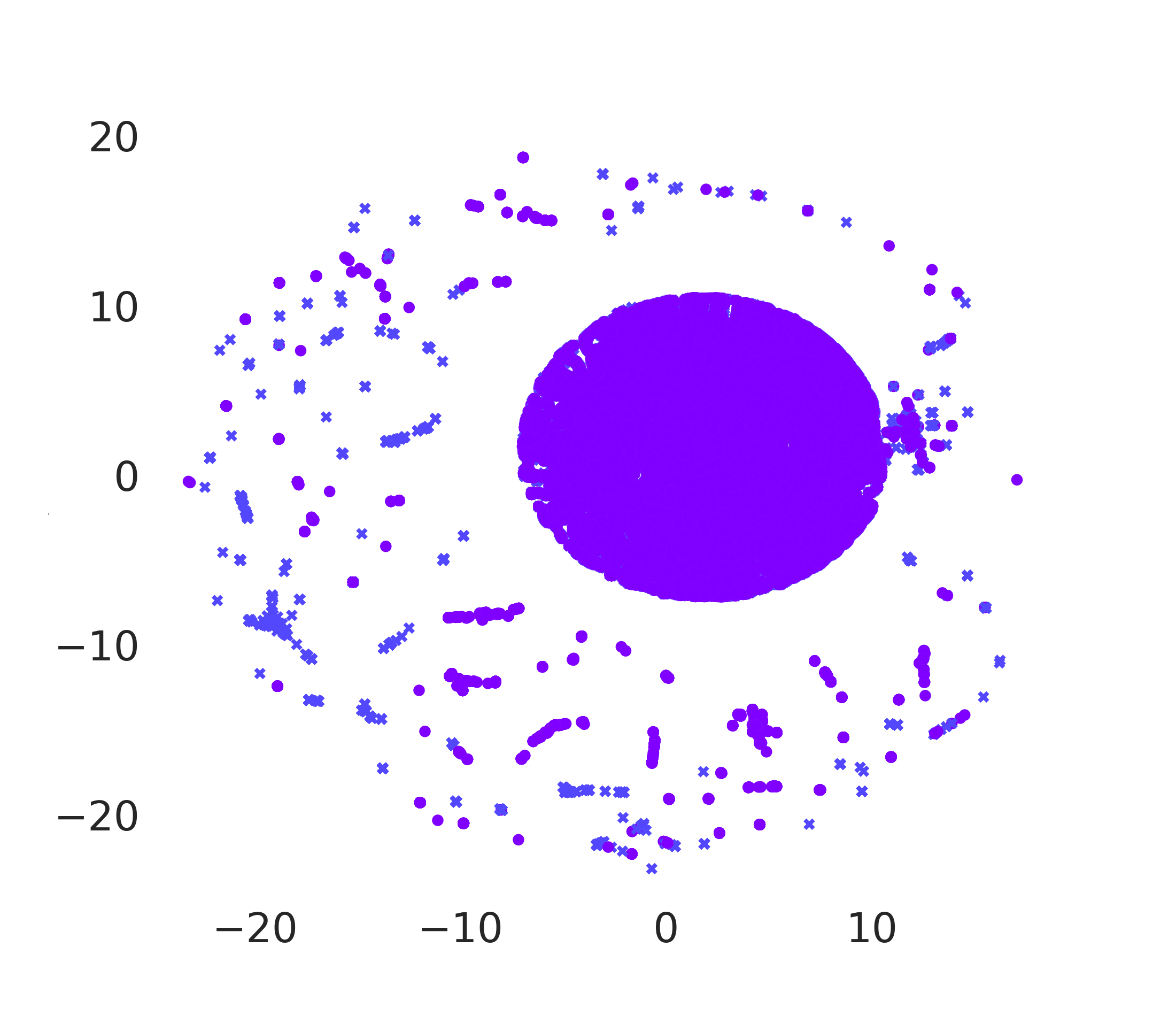}
    \caption{G-PWM2Vec}
    \label{fig_tsne_flu_gpwm2vec}
  \end{subfigure}%
  \begin{subfigure}{.15\textwidth}
    \centering
    \includegraphics[scale=0.06]{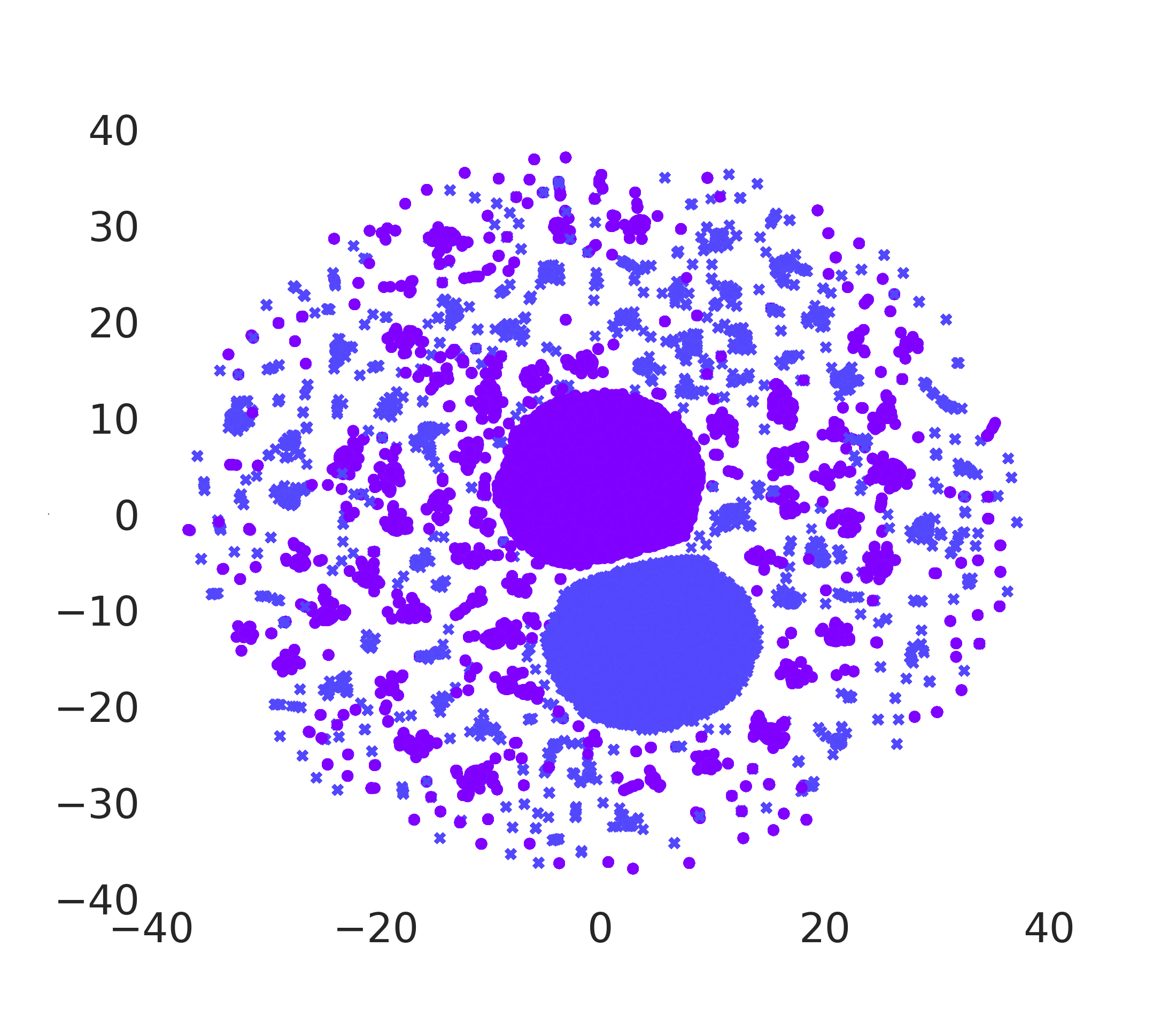}
    \caption{G-Minimizer}
    \label{fig_tsne_flu_gminimizer}
  \end{subfigure}
  \\
  \begin{subfigure}{0.45\textwidth}
    \centering
    \includegraphics[scale=0.45]{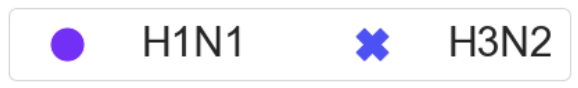}
  \end{subfigure}%
  \caption{t-SNE plots for Influenza A Virus dataset without GANs (a, b, c) and with GANs (d, e, f). The figure is best seen in color. 
 }
 \label{fig_flu_tsne}
\end{figure}

We visualize our datasets using the popular visualization technique, t-SNE~\cite{van2008visualizing}, to view the internal structure of each dataset following various embeddings. Like, the plots for the Influenza A Virus dataset are reported in Figure~\ref{fig_flu_tsne}. We can observe that for Spike2Vec and Minimizer based plots
, the addition of GANs-based features is causing two big clusters along with the small scattered clusters for each, unlike their original t-SNEs which only consist of small scattered groups. However, the PWM2Vec-based plots 
for both with GANs and without GANs show similar structures. But generally including GANs-based embeddings to the original ones can improve the t-SNE structure. 

\begin{figure}[h!]
  \centering
  \begin{subfigure}{.15\textwidth}
    \centering
    \includegraphics[scale=0.06]{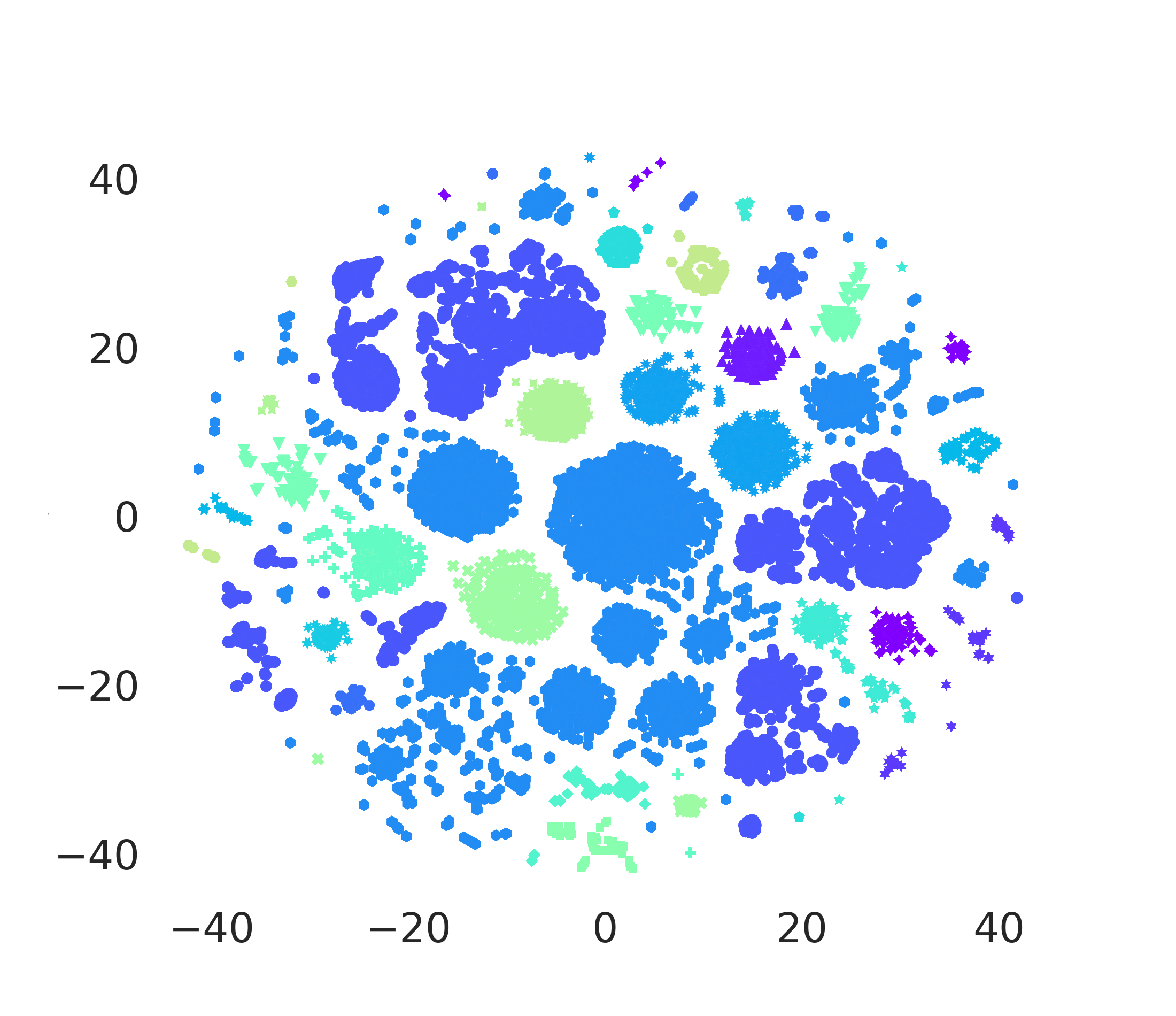}
    \caption{Spike2Vec}
    \label{fig_tsne_palmdb_spike2vec}
  \end{subfigure}%
  \begin{subfigure}{.15\textwidth}
    \centering
    \includegraphics[scale=0.06]{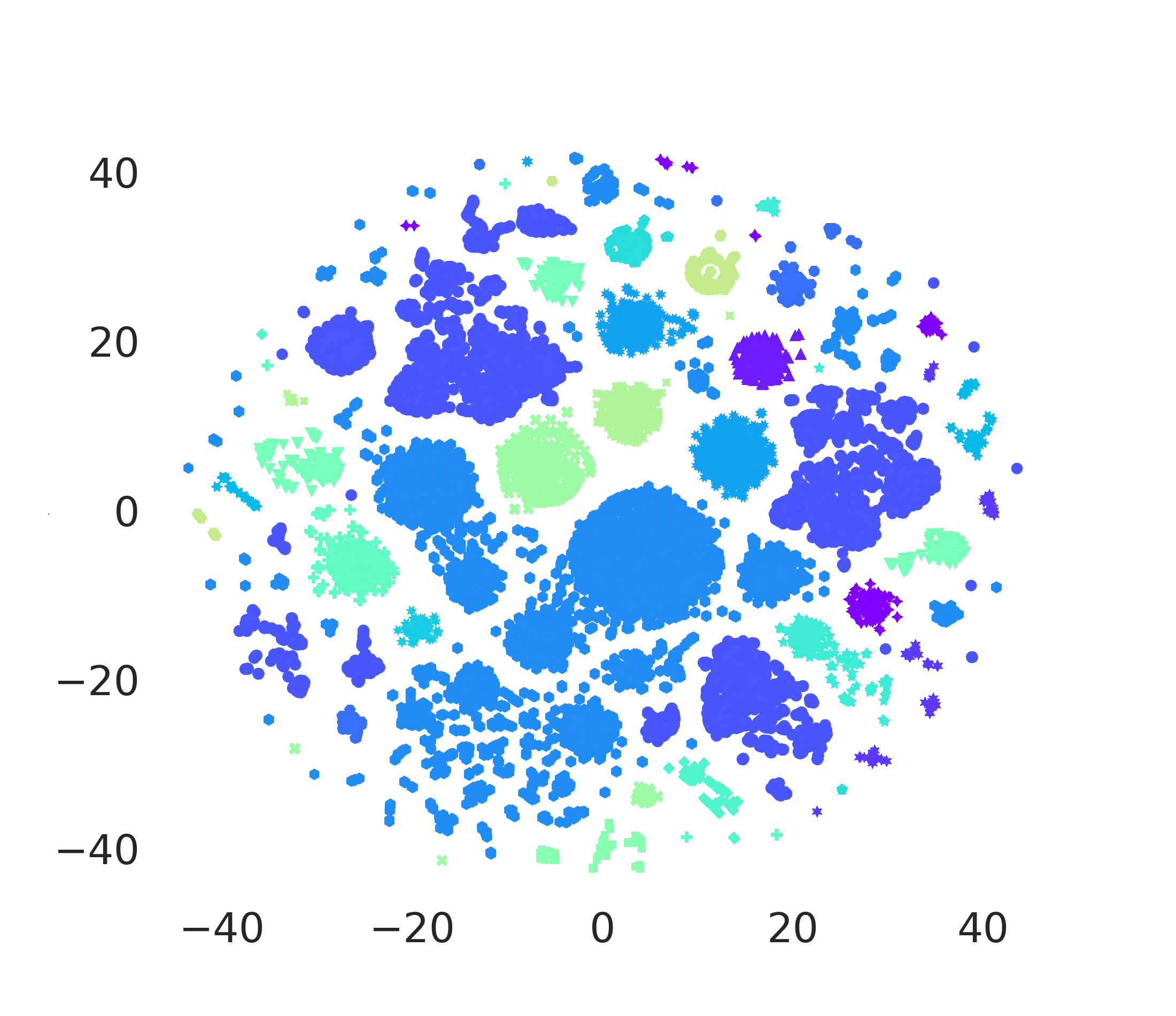}
    \caption{PWM2Vec}
    \label{fig_tsne_palmab_pwm2vec}
  \end{subfigure}%
  \begin{subfigure}{.15\textwidth}
    \centering
    \includegraphics[scale=0.06]{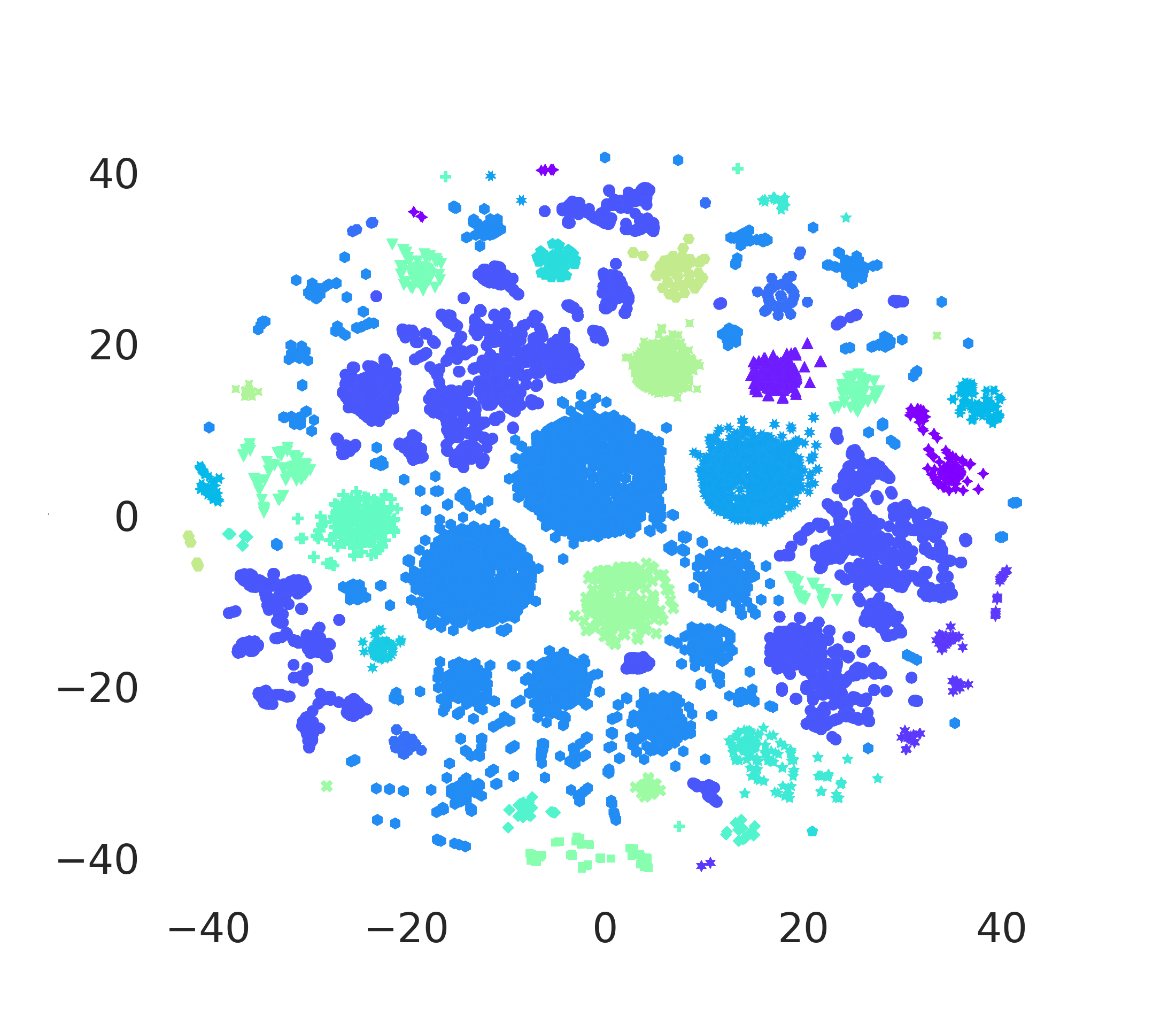}
    \caption{Minimizer}
    \label{fig_tsne_palmdb_minimizer}
  \end{subfigure}
  \\
  \begin{subfigure}{.15\textwidth}
    \centering
    \includegraphics[scale=0.06]{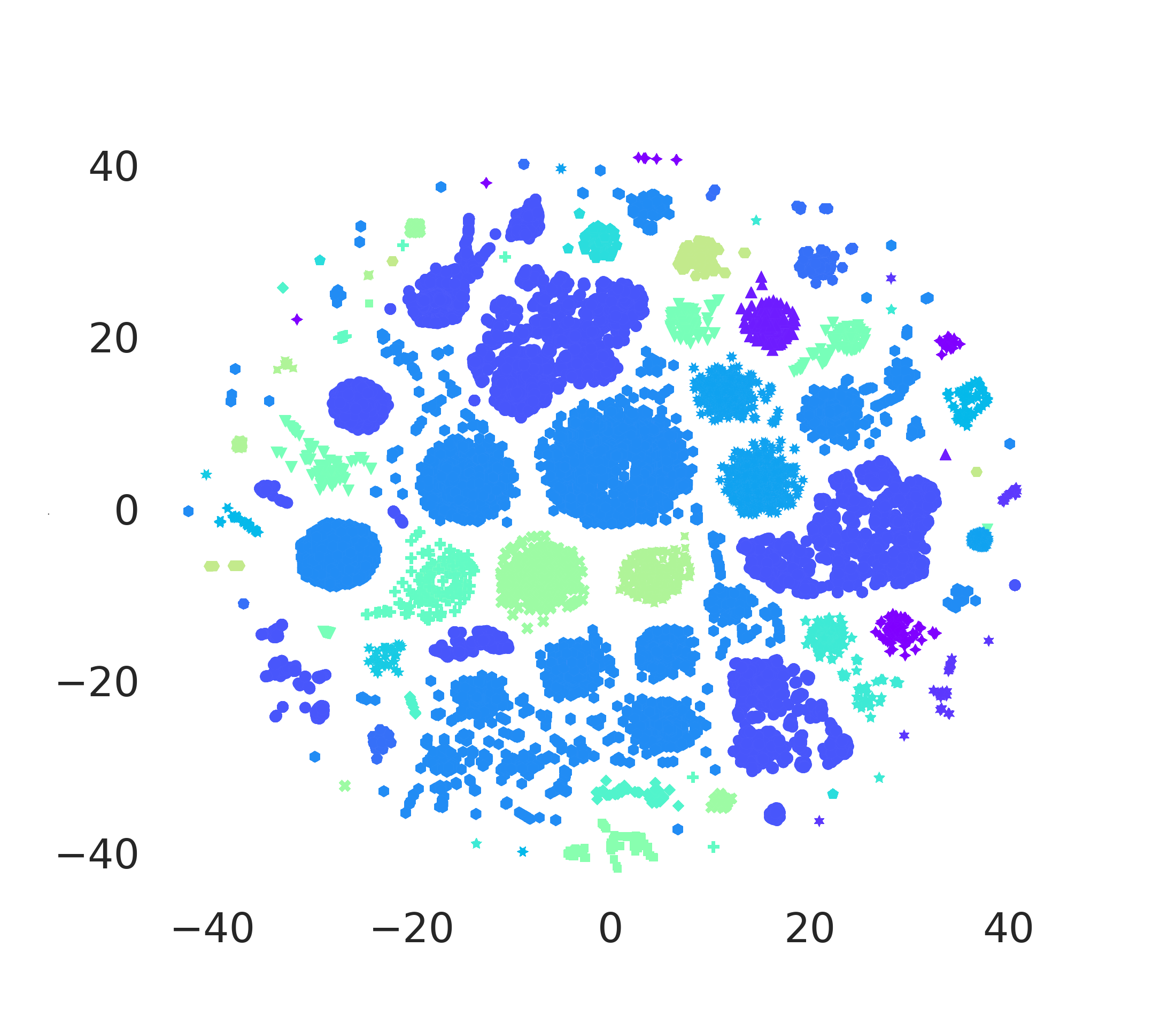}
    \caption{G-Spike2Vec}
    \label{fig_tsne_palmdb_gspike2vec}
  \end{subfigure}%
  \begin{subfigure}{.15\textwidth}
    \centering
    \includegraphics[scale=0.06]{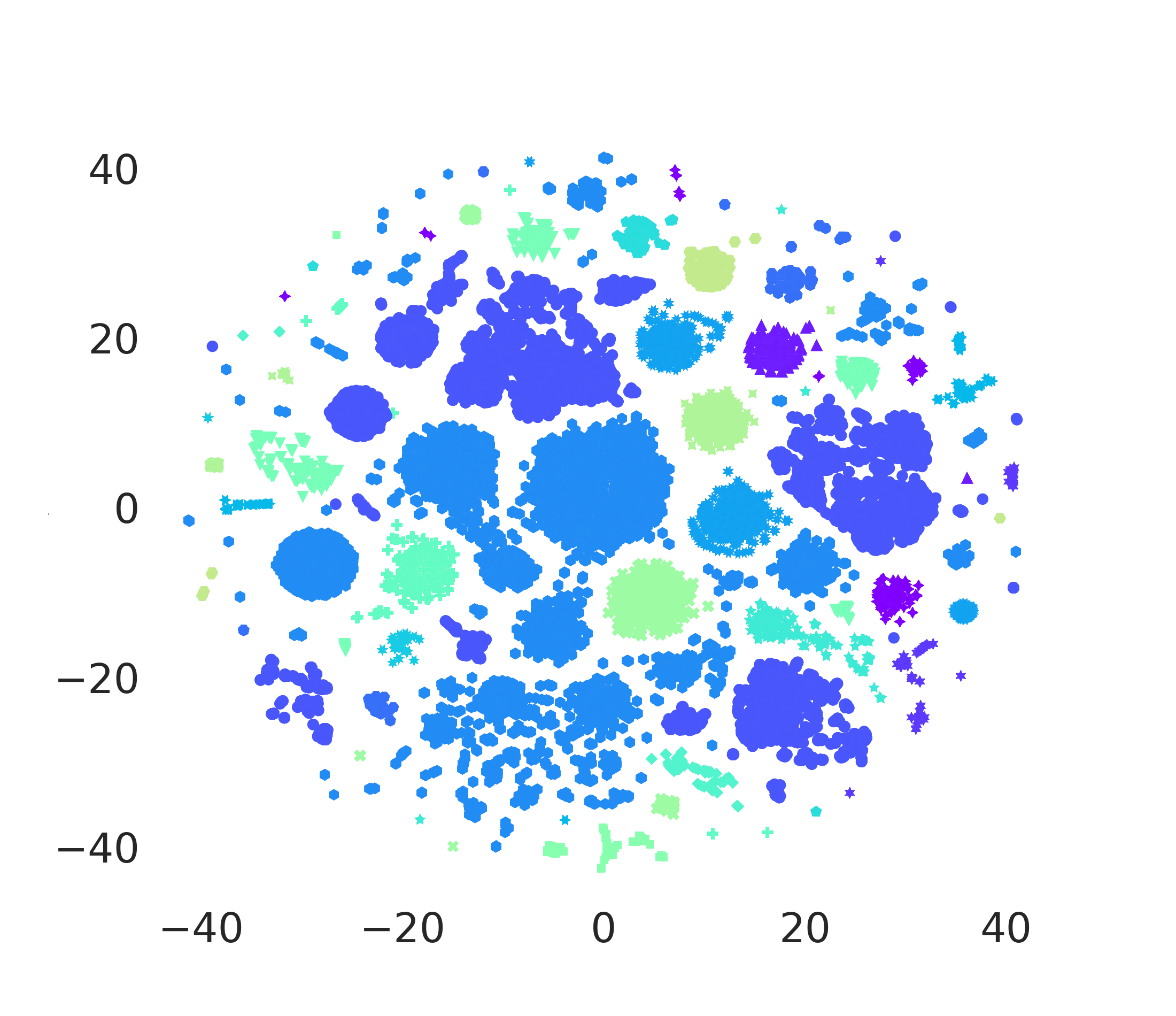}
    \caption{G-PWM2Vec}
    \label{fig_tsne_palmab_gpwm2vec}
  \end{subfigure}%
  \begin{subfigure}{.15\textwidth}
    \centering
    \includegraphics[scale=0.06]{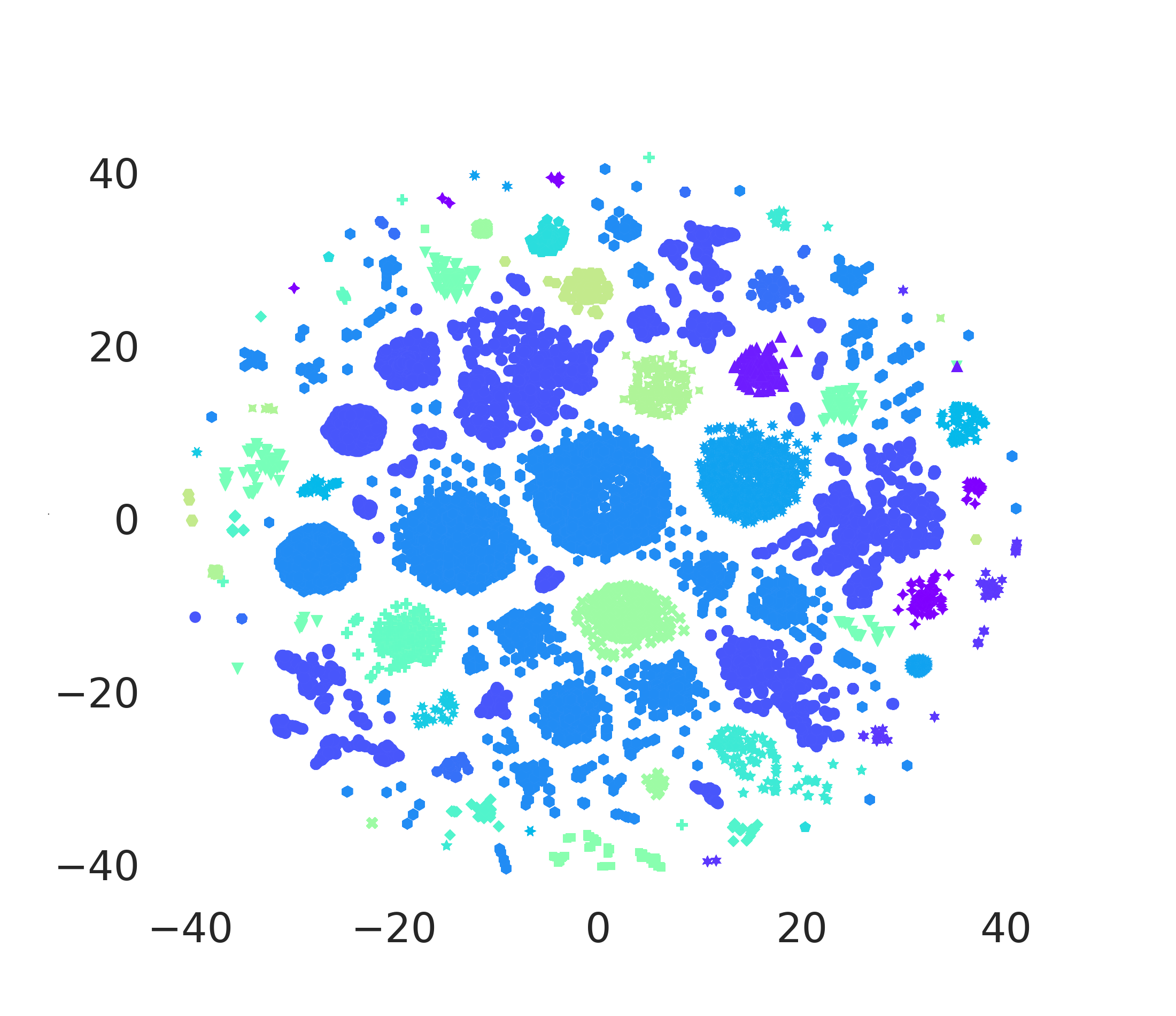}
    \caption{G-Minimizer}
    \label{fig_tsne_palmdb_gminimizer}
  \end{subfigure}
  \\
  \begin{subfigure}{0.45\textwidth}
    \centering
    \includegraphics[scale=0.34]{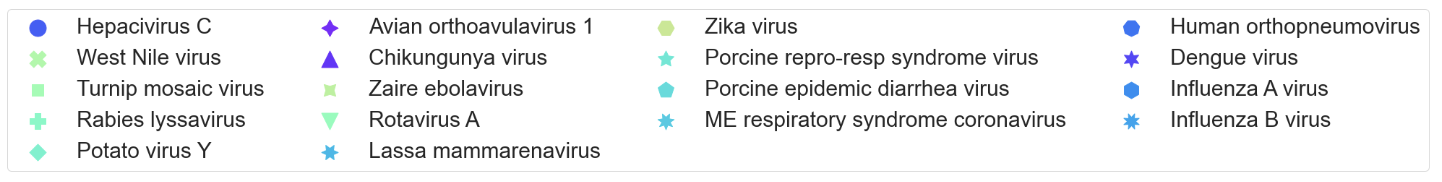}
  \end{subfigure}%
 \caption{t-SNE plots for PALMdb dataset without GANs (a, b, c), and with GANs (d, e, f). The figure is best seen in color.
 }
 \label{fig_palmdb_tsne}
\end{figure}

Moreover, the t-SNE plots for PALMdb dataset corresponding to different embeddings are shown in Figure~\ref{fig_palmdb_tsne}. 
We can observe that this dataset shows similar kinds of cluster patterns corresponding to both without GANs and with GANs based embeddings. As the original dataset is already portraying clear distinct clusters for various species, therefore adding GANs-based embedding to it is not affecting the cluster structure much. 

\begin{figure}[h!]
  \centering
  \begin{subfigure}{.15\textwidth}
    \centering
    \includegraphics[scale=0.06]{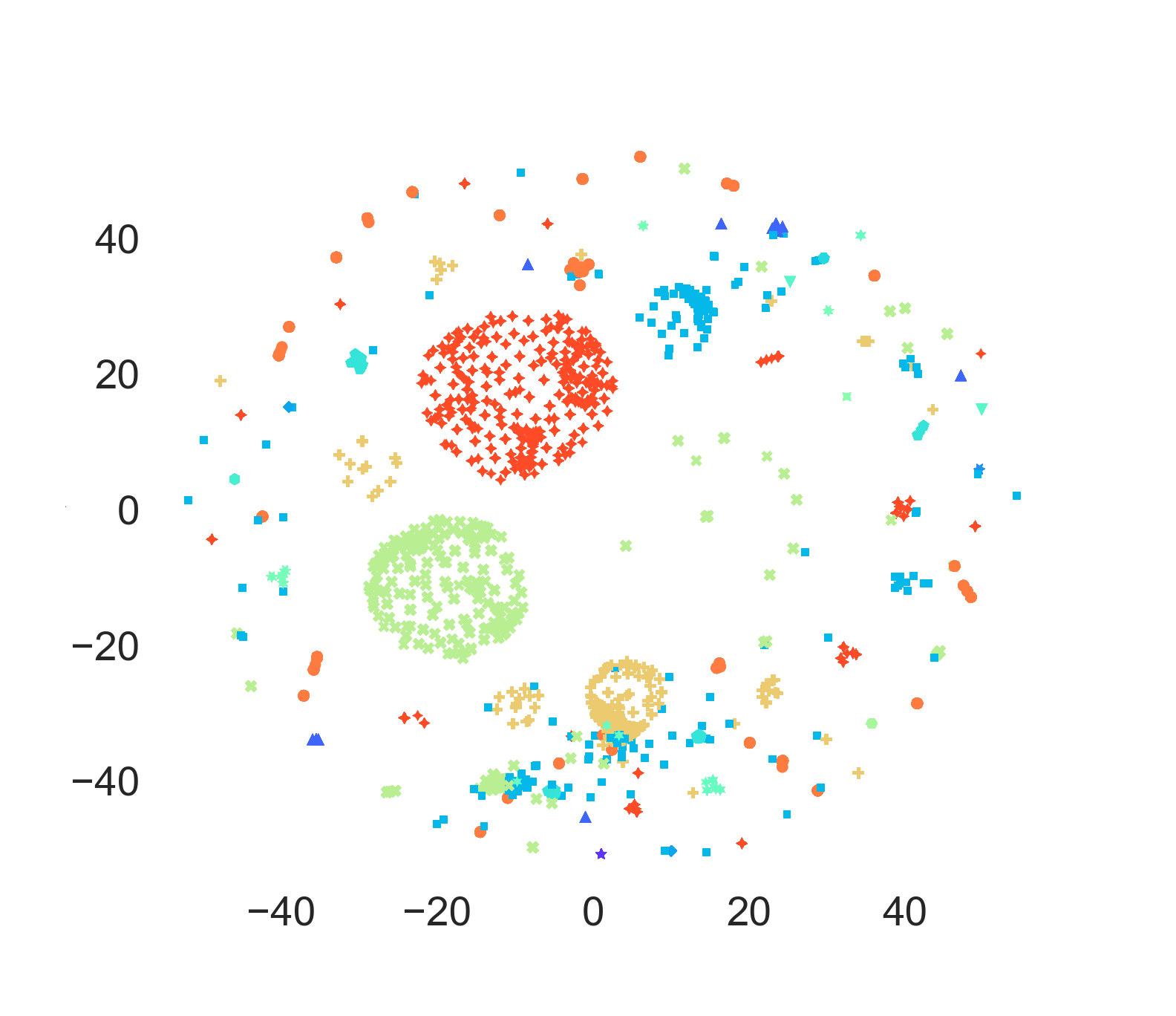}
    \caption{Spike2Vec}
    \label{fig_tsne_vdjdb_spike2vec}
  \end{subfigure}%
  \begin{subfigure}{.15\textwidth}
    \centering
    \includegraphics[scale=0.06]{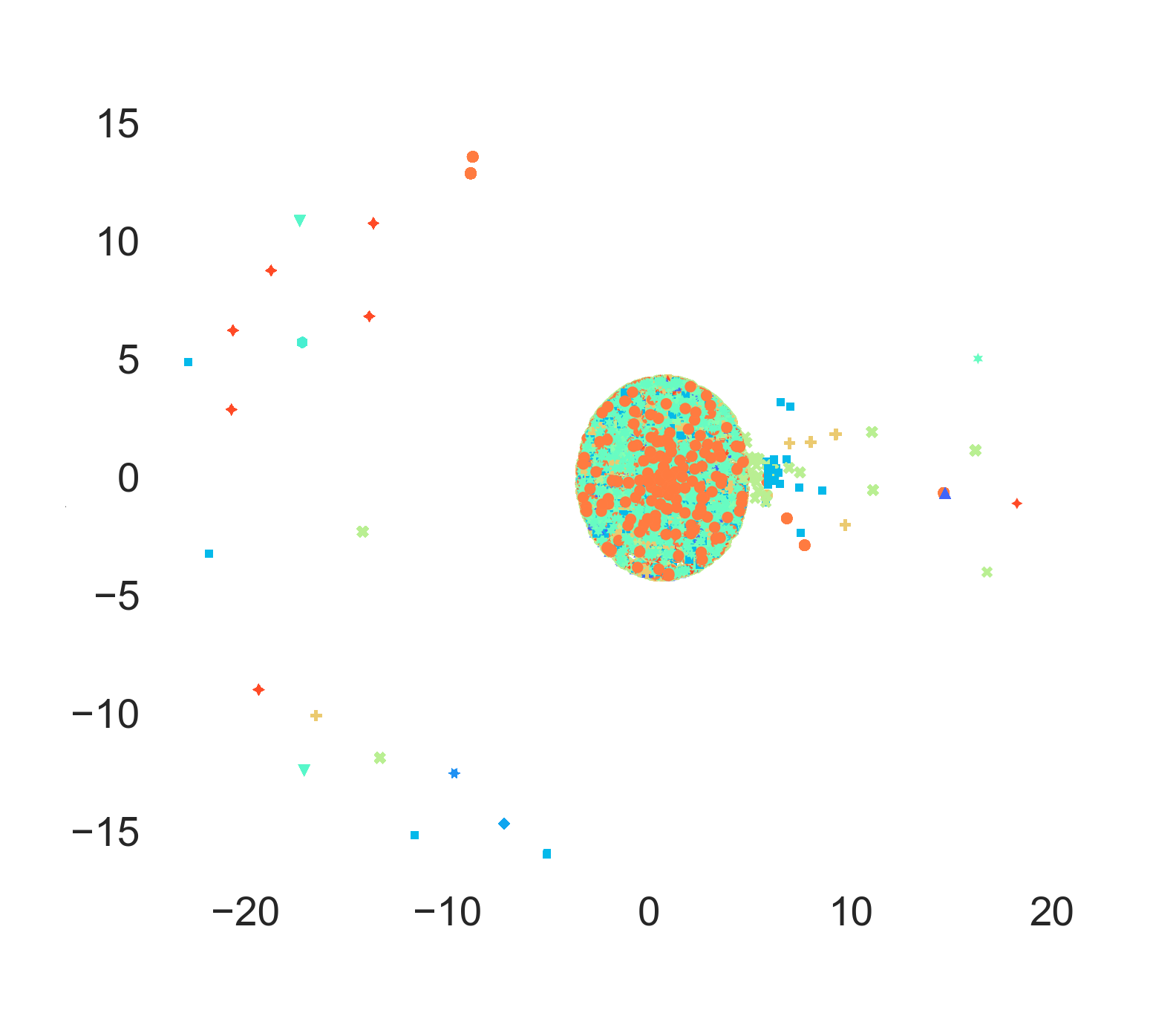}
    \caption{PWM2Vec}
    \label{fig_tsne_vdjdb_pwm2vec}
  \end{subfigure}%
  \begin{subfigure}{.15\textwidth}
    \centering
    \includegraphics[scale=0.06]{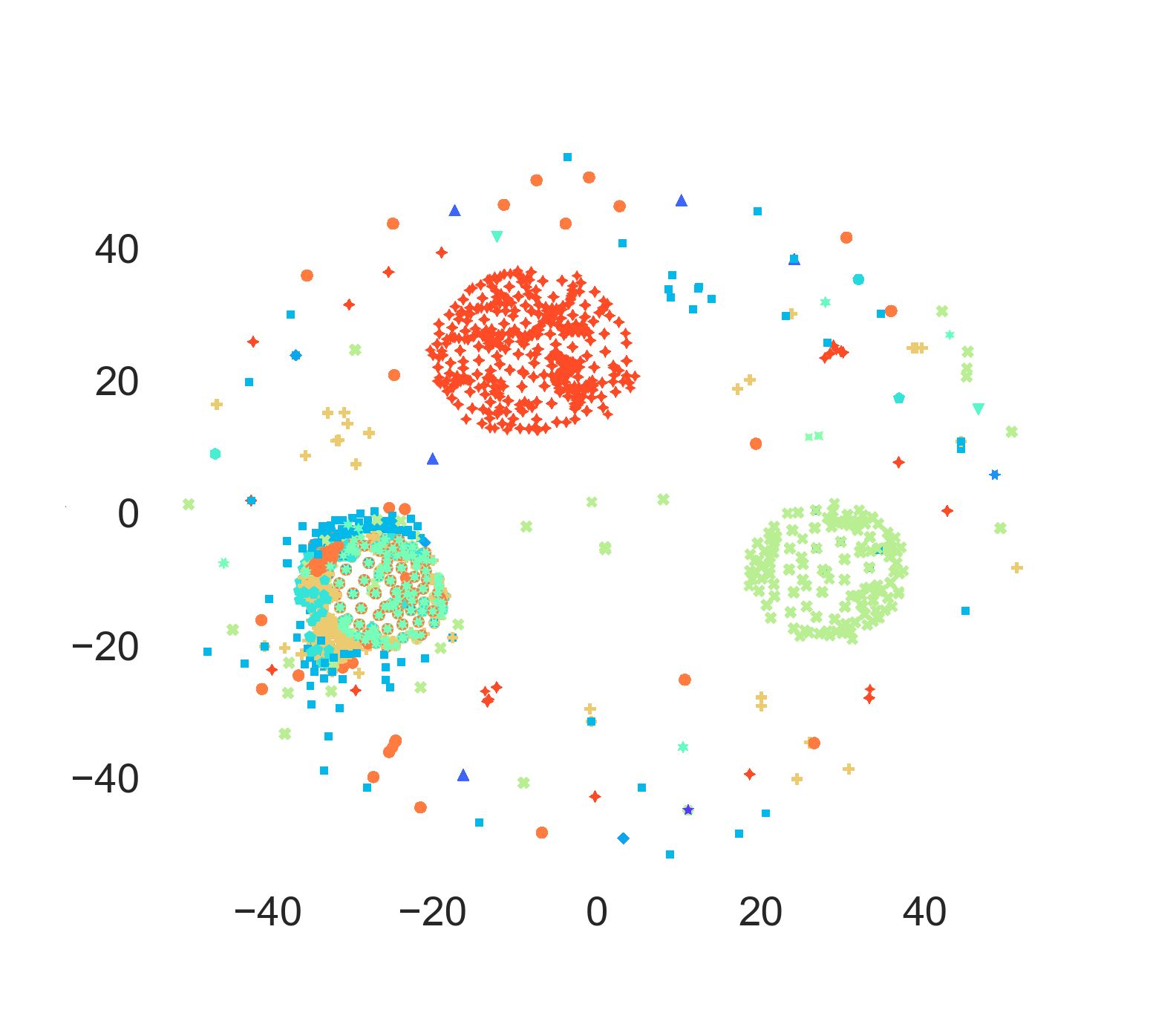}
    \caption{Minimizer}
    \label{fig_tsne_vdjdb_minimizer}
  \end{subfigure}
  \\
  \begin{subfigure}{.15\textwidth}
    \centering
    \includegraphics[scale=0.06]{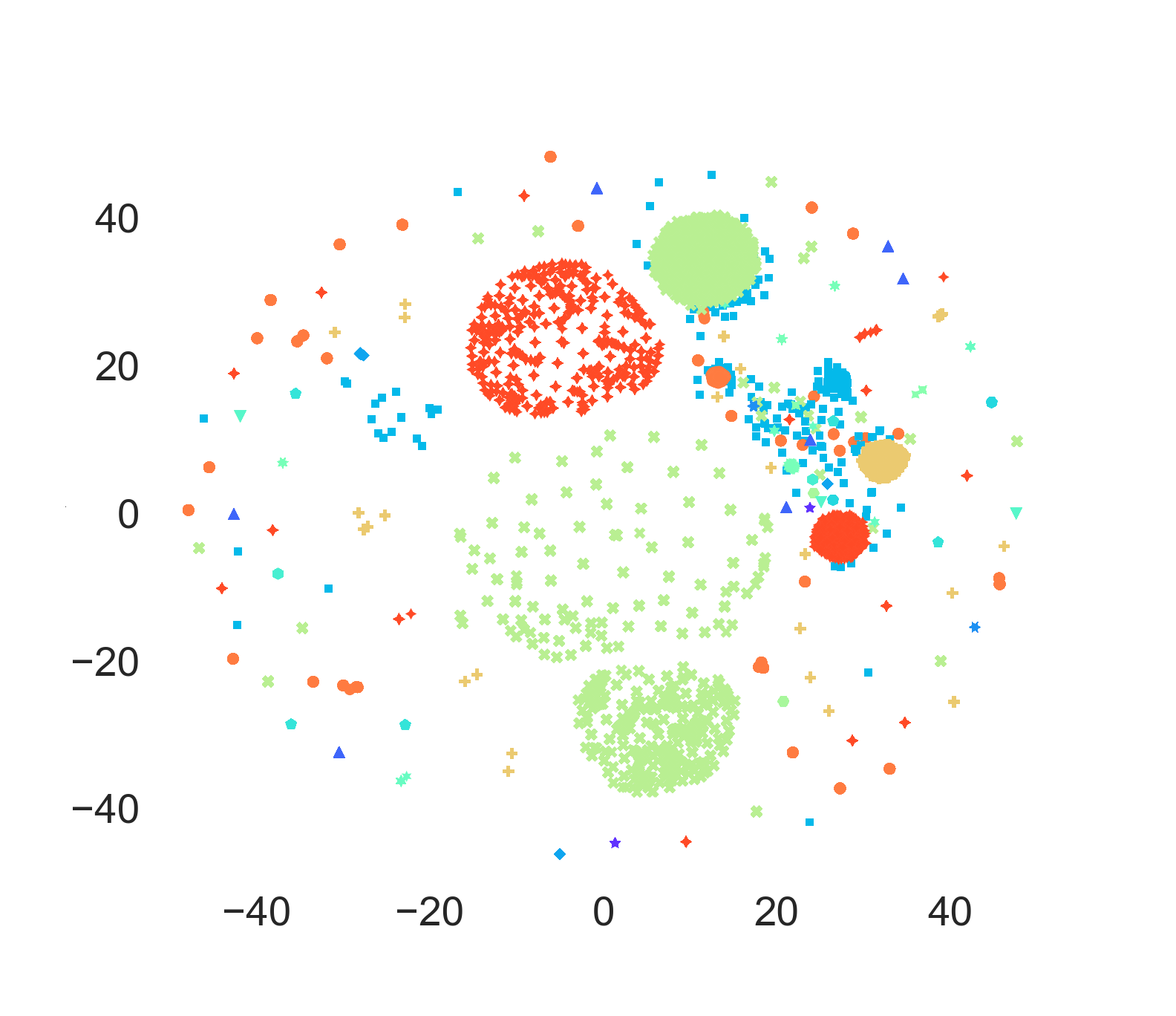}
    \caption{G-Spike2Vec}
    \label{fig_tsne_vdjdb_gspike2vec}
  \end{subfigure}%
  \begin{subfigure}{.15\textwidth}
    \centering
    \includegraphics[scale=0.06]{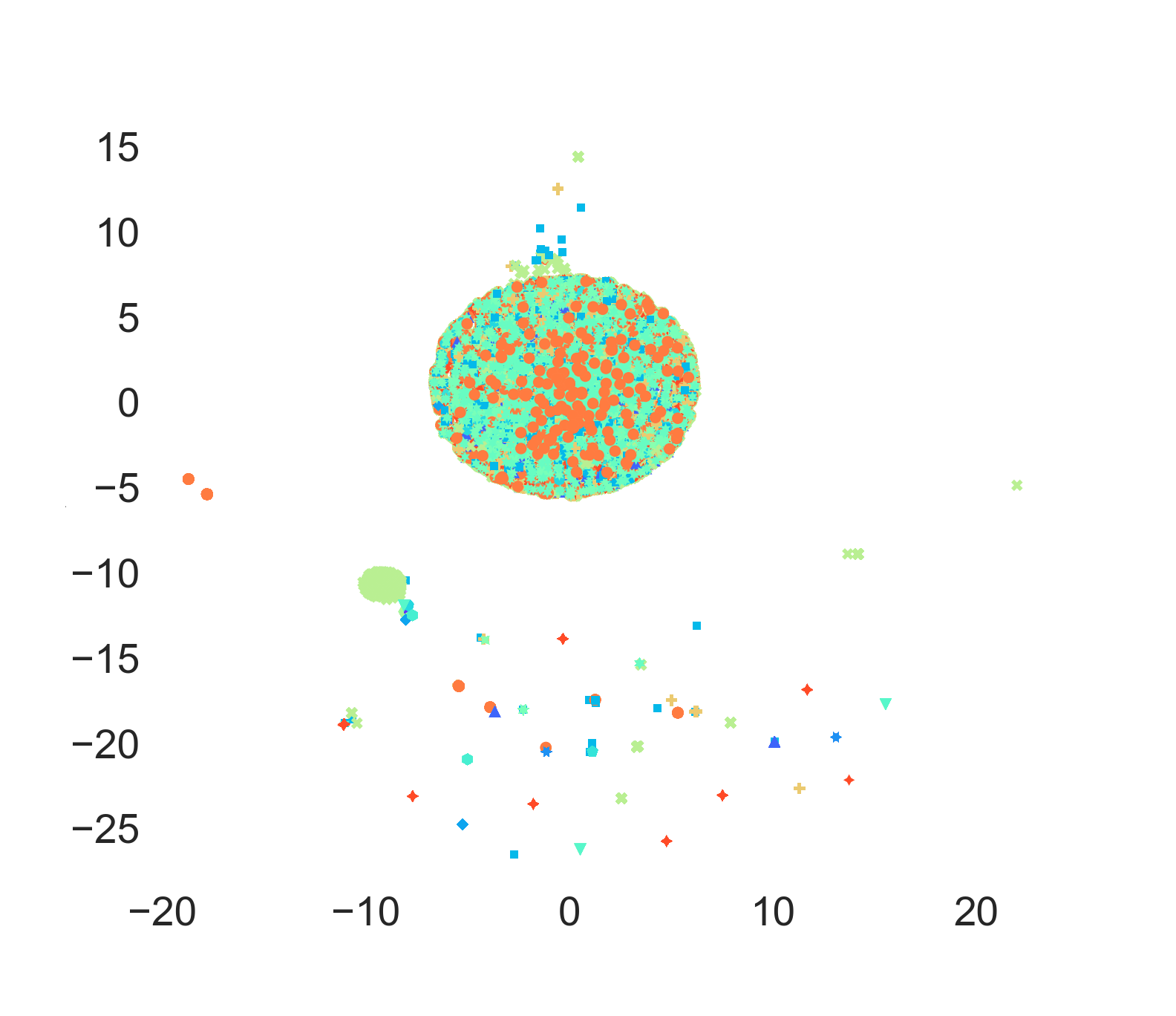}
    \caption{G-PWM2Vec}
    \label{fig_tsne_vdjdb_gpwm2vec}
  \end{subfigure}%
  \begin{subfigure}{.15\textwidth}
    \centering
    \includegraphics[scale=0.06]{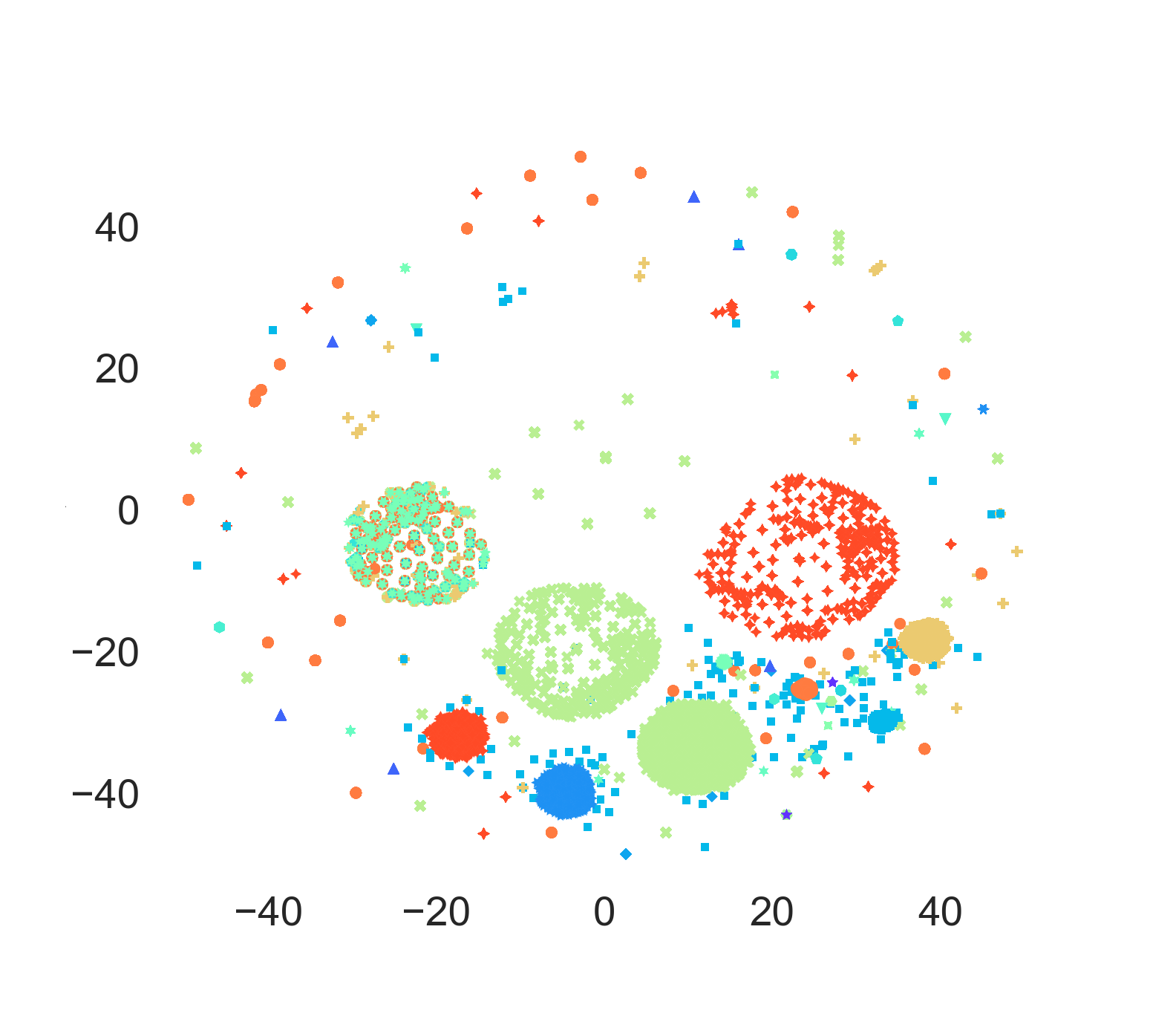}
    \caption{G-Minimizer}
    \label{fig_tsne_vdjdb_gminimizer}
  \end{subfigure}
  \\
  \begin{subfigure}{0.45\textwidth}
    \centering
    \includegraphics[scale=0.34]{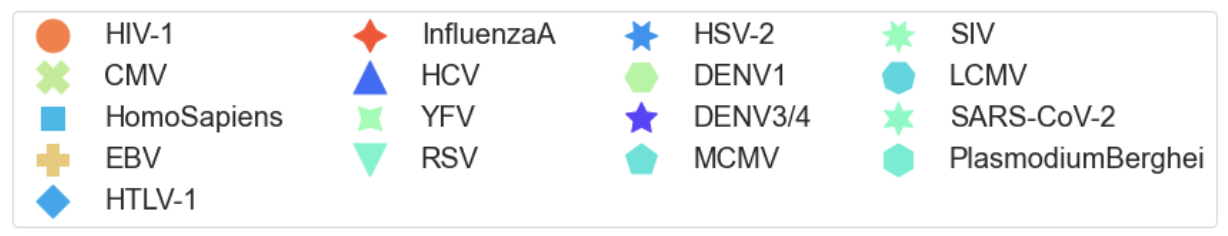}
  \end{subfigure}%
 \caption{t-SNE plots for VDjDB dataset without GANs (a, b, c) and with GANs (d, e, f). The figure is best seen in color.
 }
 \label{fig_vdjdb_tsne}
\end{figure}

Furthermore, the t-SNE plots for VDjDB dataset are given in Figure~\ref{fig_vdjdb_tsne}. 
We can observe that the addition of GAN-based features to the Minimizer-based embedding has yielded more clear and distinct clusters in the visualization. GAN-based spike2vec also portrays more clusters than the Spike2Vec one. However, the PWM2Vec shows similar patterns for both GAN-based and without GANs embeddings. Overall, it indicates that adding GANs-based features is enhancing the t-SNE cluster structures.

\begin{figure}[h!]
  \centering
  \begin{subfigure}{.15\textwidth}
    \centering
    \includegraphics[scale=0.05]{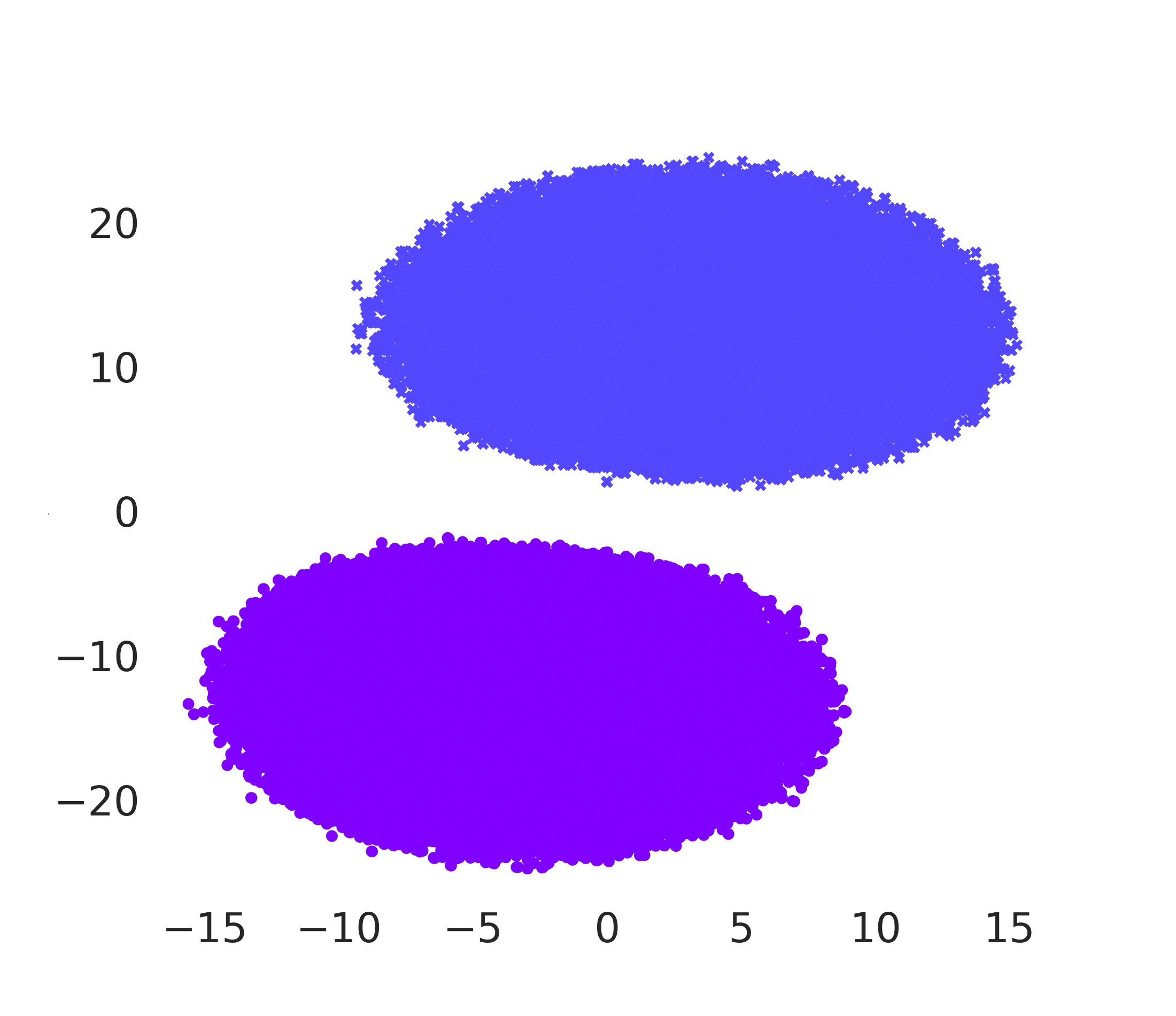}
    \caption{Influenza A Virus}
    \label{fig_tsne_onlygans_vdjdb_spike2vec}
  \end{subfigure}%
  \begin{subfigure}{.15\textwidth}
    \centering
    \includegraphics[scale=0.05]{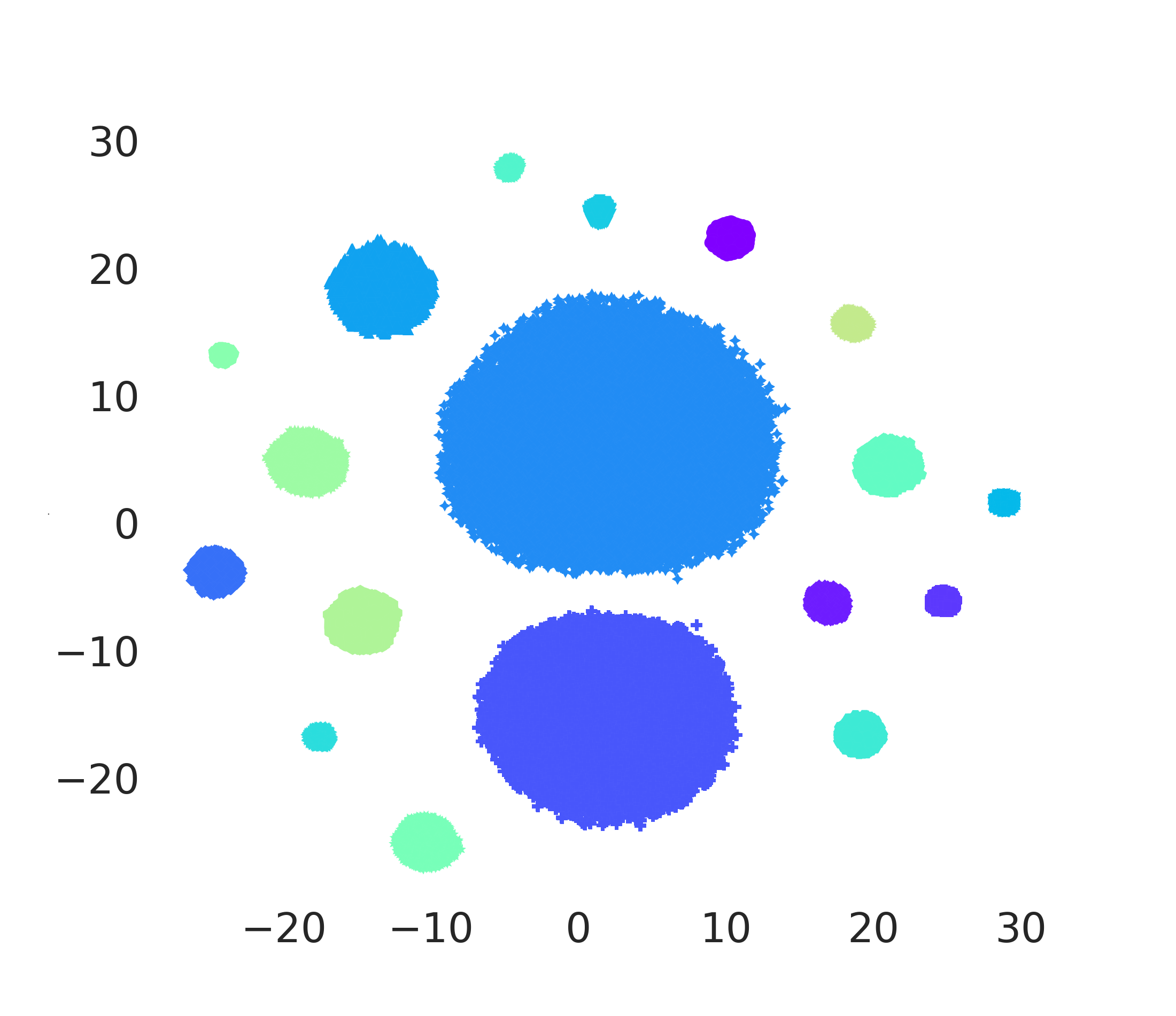}
    \caption{Palmdb}
    \label{fig_tsne_onlygans_vdjdb_pwm2vec}
  \end{subfigure}%
  \begin{subfigure}{.15\textwidth}
    \centering
    \includegraphics[scale=0.05]{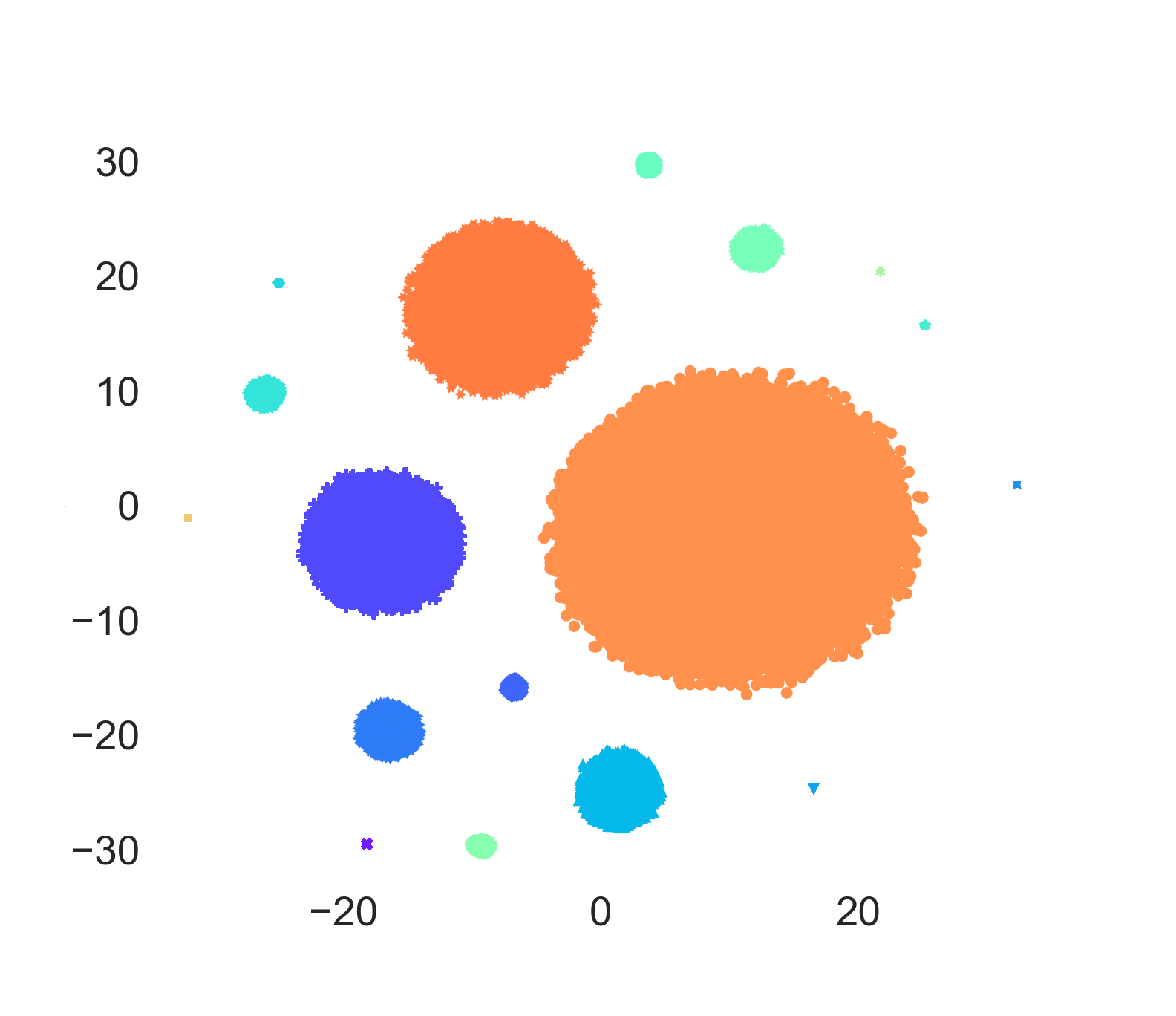}
    \caption{VDjDB}
    \label{fig_tsne_onlygans_vdjdb_minimizer}
  \end{subfigure}
  \\
 \caption{t-SNE plots of only GANs embeddings for Influenza A Virus , PALMdb, and VDjDB datasets. The figure is best seen in color. 
 }
 \label{fig_onlygans_tsne}
\end{figure}

We also investigated the t-SNE structures generated by using only the GANs-based embeddings and Figure~\ref{fig_onlygans_tsne} illustrates the results. It can be seen that for all the datasets only GAN embeddings are yielding non-overlapping distinct clusters corresponding to each group with respect to the dataset. It is because for each group the only-GAN embeddings are synthesized after training the GAN model with the original data of the respective group. 

\subsection{ML Classifiers and Evaluation Metrics}
To perform classification tasks we employ the following ML models: Naive Bayes (NB), Multilayer Perceptron (MLP), k-Nearest Neighbor (k-NN) (where $k = 3$), Random Forest (RF), Logistic Regression (LR), and Decision Tree (DT). For each classification task, the data is split into 70-30\% train-test sets using stratified sampling to preserve the original data distribution. Furthermore, our experiments are conducted by averaging the performance results of $5$ runs for each combination of dataset and classifier to get more stable results. 

We evaluated the classifiers using the following performance metrics: accuracy, precision, recall, weighted F1, F1 macro, and ROC AUC macro. Since we are doing multi-class classification in some cases, we utilized the one-vs-rest approach for computing the ROC AUC score for them. Moreover, the reason for reporting many metrics is to get more insight into the classifiers' performance, especially in the class imbalance scenario where reporting only accuracy does not provide sufficient performance information. 

\section{Results and Discussion}\label{sec_results_discussion}
This section discusses the experimental results comprehensively. The subtype classification results of the Influenza A virus dataset are given in Table~\ref{tble_results_gans}, along with the results of the PALMdb dataset species-wise classification. The antigen species-wise classification results of VDjDB data are shown in Table~\ref{tble_results_gans_2}. The reported results represent the test set results. The further details are as follows,

\subsection{Performance of Original Data}
These results illustrate the classification performance achieved corresponding to the embeddings generated by Spike2Vec, PWM2Vec, and Minimizer strategies for each dataset. We can observe that for the Influenza A Virus dataset, Spike2Vec and Minimizer are exhibiting similar performance for almost all the classifiers and are better than PWM2Vec. However, the NB model yields minimum predictive performance for all the embeddings. Similarly, the VDjDb dataset portrays similar performance for Spike2Vec and Minimizer for all evaluation metrics, while its PWM2Vec has a very low predictive performance. Moreover, all the embeddings achieve the same performance in terms of all the evaluation metrics for every classifier on the PALMdb dataset. 

\begin{table*}[h!]
    \centering
    \caption{The subtype classification results of Influenza A Virus dataset and species-wise classification results of Palmdb dataset. These results are average results over 5 runs.}
    \resizebox{0.99\textwidth}{!}{
    \begin{tabular}{@{\extracolsep{4pt}}p{0.8cm}p{1.5cm}p{0.5cm}cccccp{0.9cm}p{1.4cm}|cccccp{0.9cm}p{1.4cm}}
    \toprule
    & & & \multicolumn{7}{c}{Influenza A Virus} & \multicolumn{7}{c}{PALMdb} \\
    \cmidrule{3-10} \cmidrule{11-17}
        & Method & Algo. & Acc. $\uparrow$ & Prec. $\uparrow$ & Recall $\uparrow$ & F1 (Weig.) $\uparrow$ & F1 (Macro) $\uparrow$ & ROC AUC $\uparrow$ & Train Time (Sec.) $\downarrow$ & Acc. $\uparrow$ & Prec. $\uparrow$ & Recall $\uparrow$ & F1 (Weig.) $\uparrow$ & F1 (Macro) $\uparrow$ & ROC AUC $\uparrow$ & Train Time (Sec.) $\downarrow$ \\
        \midrule \midrule
 \multirow{18}{0.5cm}{Without GANs} & \multirow{7}{2cm}{Spike2Vec~\cite{ali2021spike2vec}}
& NB  & 0.538  & 0.673 &  0.538 &  0.382  & 0.358 &  0.503  &  96.851 & 0.999 &  0.999 &  0.999 & 0.999 &  0.999 & 0.999 &  453.961 \\
&  & MLP & 0.999 &  0.999 &   0.999 & 0.999 &  0.999 &  0.999 &  742.551 & 0.999 & 0.999 & 0.999 & 0.999 & 0.999 & 0.999 &  1446.421  \\
&  & KNN & 0.999 &  0.999 & 0.999 & 0.999 & 0.999 & 0.999 &  2689.320 & 0.999 & 0.999 & 0.999 & 0.999 & 0.999 & 0.999 &  1274.75 \\
&  & RF  & 0.999 &  0.999 &  0.999 & 0.999 &  0.999 &  0.999 &  433.459 & 0.999 & 0.999 & 0.999 & 0.999 & 0.999 & 0.999 &  166.087  \\
&  & LR  & 0.966 &  0.966 & 0.966 &  0.966  &  0.966 & 0.965  & 24.467  & 0.999 & 0.999 & 0.999 & 0.999 &  0.999 &  0.999 &  31564.898  \\
&  & DT  & 0.999 & 0.999 &  0.999 & 0.999 & 0.999 & 0.999 &  54.024  & 0.999 & 0.999 & 0.999 & 0.999 &  0.999 &  0.999 &  163.827 \\
 \cmidrule{3-10} \cmidrule{11-17}
&   \multirow{7}{2cm}{PWM2Vec~\cite{ali2022pwm2vec}}
& NB  & 0.563 & 0.745 & 0.563 & 0.435 & 0.414 & 0.530 &  60.155  & 0.999 & 0.999 & 0.999 & 0.999 & 0.999 &  0.999 &  562.922 \\
&  & MLP & 0.644 &  0.785 & 0.644 & 0.579 & 0.566 &  0.617 &  1471.086  & 0.999 &  0.999 &  0.999 & 0.999 &  0.999 &  0.999 &  1675.896 \\
&  & KNN & 0.644 & 0.785 & 0.644 & 0.579 & 0.566 & 0.617 & 2665.538  & 0.999 &  0.999 & 0.999 & 0.998 &  0.999 &  0.999 &  1514.240 \\
&  & RF  & 0.644 & 0.785 & 0.644 & 0.579 & 0.566 & 0.618 & 1514.979  & 0.999 &  0.999 & 0.999 & 0.999 & 0.999 & 0.999 &  284.450 \\
&  & LR  & 0.644 & 0.784 & 0.644 & 0.579 & 0.565 & 0.617 &  388.235  & 0.999 &  0.999 & 0.999 & 0.999 & 0.999 & 0.999 &  41029.833 \\
&  & DT  & 0.644 & 0.785 & 0.644 & 0.579 & 0.566 & 0.617 &  78.525  & 0.999 &  0.999 &  0.999 & 0.999 & 0.999 & 0.999 &  233.533 \\
  \cmidrule{3-10} \cmidrule{11-17}
&   \multirow{7}{2cm}{Minimizer}
& NB  & 0.679 &  0.682 & 0.679 & 0.673 &  0.669 &  0.670 & 57.469  & 0.999 &  0.999 & 0.999 & 0.999 &  0.999 &  0.999 &  474.482 \\
&  & MLP & 0.998 &  0.998 & 0.998 & 0.998 &  0.998 &  0.998 &  1864.844  & 0.999 & 0.999 & 0.999 & 0.999 & 0.999 & 0.999 &  3958.188 \\
&  & KNN & 0.999 & 0.999 & 0.999 & 0.999 & 0.999 & 0.999 &  2818.292  & 0.999 &  0.999 &  0.999 & 0.999 &  0.999 &  0.999 &  1357.673 \\
&  & RF  & 0.999 &  0.999 &  0.999 & 0.999 &  0.999 &  0.999 &  1039.824  & 0.999 & 0.999 & 0.999 & 0.999 & 0.999 & 0.999 &  399.507 \\
&  & LR  & 0.719 & 0.719 & 0.719 & 0.719 &  0.718 &  0.718 &  186.522  & 0.999 & 0.999 & 0.999 & 0.999 & 0.999 & 0.999 &  7270.111 \\
&  & DT  & 0.999 &  0.999 & 0.999 & 0.999 & 0.999 & 0.999 &  72.510  & 0.999 & 0.999 & 0.999 & 0.999 & 0.999 & 0.999 &  223.215 \\
\midrule
 \multirow{18}{0.5cm}{With GANs} & \multirow{7}{2cm}{Spike2Vec~\cite{ali2021spike2vec}}
&  NB  & 0.538 & 0.681 & 0.538 & 0.380 & 0.355 & 0.502 & 138.179  & 0.999 &  0.999 &  0.999 & 0.999 &  0.999 &  0.999 &  197.033 \\
&  & MLP & 0.992 & 0.992 & 0.992 & 0.992 & 0.992 & 0.992 &  1604.287  & 0.999 &  0.999 & 0.999 & 0.999 & 0.999 &  0.999 &  491.182 \\
&  & KNN & 0.999 & 0.999 &  0.999 & 0.999 &  0.999 & 0.999 & 3546.211  & 0.999 & 0.999 & 0.999 & 0.999 & 0.999 & 0.999 &  689.672 \\
&  & RF  & 0.999 & 0.999 & 0.999 & 0.999 & 0.999 & 0.999 & 784.393  & 0.999 & 0.999 & 0.999 & 0.999 &  0.999 & 0.999 &  243.646 \\
&  & LR  & 0.957 & 0.957 & 0.957 & 0.957 & 0.957 & 0.957 & 6810.398 & 0.999 & 0.999 & 0.999 & 0.999 &  0.999 & 0.999 &  2643.646 \\
&  & DT  & 0.999 & 0.999 & 0.999 & 0.999 & 0.999 & 0.999 & 365.332  & 0.999 & 0.999 & 0.999 & 0.999 & 0.999 &  0.999 &   396.362 \\
 \cmidrule{3-10} \cmidrule{11-17}
&   \multirow{7}{2cm}{PWM2Vec~\cite{ali2022pwm2vec}}
& NB  & 0.565 & 0.748 &  0.565 & 0.437 & 0.416 & 0.532 &  107.617  & 0.999 & 0.999 & 0.999 & 0.999 & 0.999 &  0.999 & 569.510 \\
&  & MLP & 0.644 & 0.784 &  0.644 & 0.579 & 0.566 & 0.617 & 1817.859  & 0.999 & 0.999 & 0.999 & 0.999 & 0.999 & 0.999 &  1337.920 \\
&  & KNN & 0.646 & 0.785 & 0.646 & 0.581 & 0.568 & 0.619 &  2965.701  & 0.999 & 0.999 & 0.999 & 0.999 & 0.999 & 0.999 &  1524.009 \\
&  & RF  & 0.646 &  0.786 &  0.646 & 0.582 &  0.569 &  0.619 &  1837.425  & 0.999 & 0.999 & 0.999 & 0.999 & 0.999 & 0.999 & 1802.577 \\
&  & LR  & 0.632 & 0.793 & 0.632 & 0.589 & 0.597 & 0.657 &  10273.672  & 0.999 & 0.999 & 0.999 & 0.999 & 0.999 & 0.999 &  3549.095 \\
&  & DT  & 0.646 & 0.786 & 0.646 & 0.581 & 0.568 & 0.619 & 1264.188  & 0.999 & 0.999 & 0.999 & 0.999 & 0.999 & 0.999 & 2580.831 \\
  \cmidrule{3-10} \cmidrule{11-17}
&   \multirow{7}{2cm}{Minimizer}
& NB  & 0.611 & 0.726 & 0.611 & 0.534 & 0.520 & 0.584 &  127.058  & 0.999 & 0.999 & 0.999 & 0.999 & 0.999 & 0.999 & 669.513 \\
&  & MLP & 0.976 &  0.976 & 0.976 & 0.976 & 0.976 & 0.976 &  825.868  & 0.999 &  0.999 &  0.999 & 0.999 & 0.999 & 0.999 &  1231.650 \\
&  & KNN & 0.999 & 0.999 & 0.999 & 0.999 &  0.999 &  0.999 &  3163.325  & 0.999 &  0.999 &  0.999 & 0.999 & 0.999 & 0.999 &  1484.555 \\
&  & RF  & 0.999 & 0.999 & 0.999 & 0.999 & 0.999 & 0.999 &  1557.065  & 0.999 &  0.999 & 0.999 & 0.999 & 0.999 & 0.999 &  1699.503 \\
&  & LR  & 0.711 &  0.712 &  0.711 & 0.711 &  0.710 &  0.711 &  2179.485  & 0.999 &  0.999 & 0.999 & 0.999 & 0.999 & 0.999 &  3482.345 \\
&  & DT  & 0.999 &  0.999 &  0.999 & 0.999 &  0.999 &  0.999 &  481.232  & 0.999 &  0.999 & 0.999 & 0.999 & 0.999 &  0.999 &  2700.860 \\
\midrule
 \multirow{18}{1cm}{Only GANs For Training} & \multirow{7}{2cm}{Spike2Vec~\cite{ali2021spike2vec}}
& NB  & 0.443 & 0.318 &  0.443 & 0.296 & 0.317 &  0.476 & 69.293  & 0.056 & 0.005 & 0.056 & 0.009 & 0.014 & 0.523 &  172.517 \\
&  & MLP & 0.499 & 0.506 &  0.499 & 0.498 &  0.499 &  0.503 &  279.364  & 0.104 & 0.260 & 0.104 & 0.148 & 0.039 & 0.486 & 264.306 \\
&  & KNN & 0.586 & 0.623 & 0.586 & 0.523 & 0.510 &  0.561 &  4088.144  & 0.126 & 0.242 & 0.126 & 0.156 & 0.123 & 0.533 &  263.101 \\
&  & RF  & 0.464 &  0.215 & 0.464 & 0.294 & 0.317 & 0.500 &  386.409  & 0.011 & 0.000 & 0.011 & 0.000 &  0.001 &  0.500 &  8451.755 \\
&  & LR  & 0.523 & 0.523 & 0.523 & 0.523 & 0.520 & 0.520 &  469.512  & 0.001 & 0.000 & 0.001 & 0.000 &  0.001 &  0.500 &  1481.505 \\
&  & DT  & 0.535 & 0.286 &  0.535 &  0.373 & 0.348 &  0.500 &  308.698  & 0.042 & 0.001 & 0.042 & 0.003 &   0.004 &  0.499 & 2764.815 \\
 \cmidrule{3-10} \cmidrule{11-17}
&   \multirow{7}{2cm}{PWM2Vec~\cite{ali2022pwm2vec}}
& NB  & 0.468 &  0.508 &  0.468 & 0.331 &  0.351 &  0.500 &   60.008  & 0.034 & 0.004 & 0.034 & 0.003 &  0.004 & 0.499 &  370.330 \\
&  & MLP & 0.471 & 0.503 &  0.471 & 0.369 &  0.385 &  0.500 &  333.503  & 0.400 & 0.335 &  0.400 & 0.355 & 0.080 & 0.534 &  577.936 \\
&  & KNN & 0.520 &  0.575 & 0.520 & 0.470 & 0.480 &  0.542 &  4565.427  & 0.061 &  0.213 &  0.061 & 0.089 & 0.059 & 0.496 &  2475.871 \\
&  & RF  & 0.535 &  0.286 & 0.535 & 0.372 & 0.348 &  0.500 &  746.999  & 0.034 & 0.001 & 0.034 & 0.002 & 0.003 &  0.500 &  10880.182 \\
&  & LR  & 0.534 & 0.603 & 0.534 & 0.482 &  0.492 &  0.557 &  975.877  & 0.001 &  0.012 &  0.001 & 0.009 & 0.009 & 0.490 &  278.851\\
&  & DT  & 0.535 &  0.286 &  0.535 & 0.372 & 0.348 &  0.500 & 500.541  & 0.022 & 0.001 & 0.022 & 0.032 & 0.013 &  0.500 &  3078.085 \\
  \cmidrule{3-10} \cmidrule{11-17}
&   \multirow{7}{2cm}{Minimizer}
& NB  & 0.523 & 0.529 & 0.523 & 0.523 & 0.523 & 0.526 & 65.955  & 0.062 & 0.194 & 0.062 & 0.048 & 0.055 & 0.525 & 497.483 \\
&  & MLP & 0.477 & 0.495 & 0.477 & 0.447 & 0.455 & 0.494 & 499.569  & 0.005 & 0.003 & 0.005 & 0.003 & 0.008 & 0.475 & 707.236 \\
&  & KNN & 0.539 & 0.538 & 0.539 & 0.538 & 0.535 & 0.536 &  5211.216  & 0.177 & 0.155 & 0.177 & 0.148 & 0.058 & 0.522 &  3116.525 \\
&  & RF  & 0.535 & 0.287 &  0.535 & 0.373 & 0.348 & 0.499 &  624.564  & 0.034 &  0.001 & 0.034 & 0.002 & 0.003 &  0.500 &  10349.430 \\
&  & LR  & 0.548 & 0.548 & 0.548 & 0.548 & 0.546 & 0.546 &  771.273  & 0.201 & 0.120 & 0.201 & 0.228 & 0.102 & 0.501 &  3234.386 \\
&  & DT  & 0.464 & 0.215 &  0.464 & 0.294 & 0.317 & 0.500 &  576.693  & 0.003 &  0.002 & 0.003 & 0.002 & 0.003 &  0.500 &  346.660 \\
         \bottomrule
         \end{tabular}
    }
    \label{tble_results_gans}
\end{table*}

\subsection{Performance of Original Data with GANs}
To view the impact of GAN-based data on the predictive performance for all the datasets, we evaluate the performance using the original embeddings with GAN-based synthetic data added to them respectively. This GANs-based data is used to train the classifiers, while only the original data is used as test data. 
To generate the GAN data corresponding to an embedding generation method, the GAN model is trained with the original embeddings first and then new data is synthesized for that embedding. Every label of the embedding will have a different count of synthetic data depending on its count in the original embedding data. We use $10\%$ of the original counts to generate the synthetic embeddings respectively. 

\begin{table}[h!]
    \centering
    \caption{The antigen species-wise classification results of VDjDB dataset. These results are average values over 5 runs. 
    }
    \resizebox{0.49\textwidth}{!}{
    \begin{tabular}{@{\extracolsep{4pt}}p{0.8cm}p{1.5cm}p{0.5cm}cccccp{0.9cm}p{1.4cm}}
    \toprule
        & Method & Algo. & Acc. $\uparrow$ & Prec. $\uparrow$ & Recall $\uparrow$ & F1 (Weig.) $\uparrow$ & F1 (Macro) $\uparrow$ & ROC AUC $\uparrow$ & Train Time (Sec.) $\downarrow$\\
        \midrule \midrule
 \multirow{18}{0.5cm}{Without GANs} & \multirow{7}{2cm}{Spike2Vec~\cite{ali2021spike2vec}}
& NB  & 0.999 & 0.999 & 0.999 & 0.999 & 0.999 & 0.999 &  87.948 \\
&  & MLP & 0.999 & 0.999 & 0.999 & 0.999 & 0.999 & 0.999 &  689.357 \\
&  & KNN & 0.998 & 0.998 & 0.998 & 0.998 &  0.998 &  0.999 & 167.426 \\
&  & RF  & 0.999 & 0.999 & 0.999 & 0.999 &  0.999 &  0.999 &  152.581 \\
&  & LR  & 0.999 & 0.999 & 0.999 & 0.999 &  0.999 &  0.999 &  882.695 \\
&  & DT  & 0.999 & 0.999 & 0.999 & 0.999 &  0.999 &  0.999 &  43.314 \\
 \cmidrule{3-10}
&   \multirow{7}{2cm}{PWM2Vec~\cite{ali2022pwm2vec}}
& NB  & 0.179 & 0.926 & 0.179 & 0.250 & 0.305 & 0.634 &  84.292 \\
&  & MLP & 0.525 & 0.685 & 0.525 & 0.399 & 0.315 & 0.626 &  1216.913 \\
&  & KNN & 0.525 & 0.689 & 0.525 & 0.399 & 0.320 & 0.626 &  248.660 \\
&  & RF  & 0.525 & 0.690 & 0.525 & 0.400 & 0.320 & 0.626 & 736.583 \\
&  & LR  & 0.525 & 0.681 & 0.525 & 0.400 & 0.320 &  0.626 &  299.575 \\
&  & DT  & 0.525 & 0.690 & 0.525 & 0.400 &  0.320 & 0.626 &  39.697 \\
  \cmidrule{3-10} 
&   \multirow{7}{2cm}{Minimizer}
& NB  & 0.930 & 0.972 & 0.930 & 0.940 & 0.838 & 0.935 &  98.159 \\
&  & MLP & 0.934 & 0.952 &   0.934 & 0.928 & 0.782 & 0.882 &  1253.018 \\
&  & KNN & 0.951 & 0.961 & 0.951 & 0.947 & 0.849 & 0.925 &  172.851 \\
&  & RF  & 0.953 & 0.962 & 0.953 & 0.948 & 0.847 & 0.927 &  468.139 \\
&  & LR  & 0.952 &  0.961 & 0.952 & 0.948 & 0.847 &  0.926 &  203.061 \\
&  & DT  & 0.952 &  0.962 &  0.952 & 0.948 & 0.847 & 0.926 &  25.392 \\
\midrule
 \multirow{18}{0.5cm}{With GANs} & \multirow{7}{2cm}{Spike2Vec~\cite{ali2021spike2vec}}
& NB  & 0.999 &  0.999 & 0.999 & 0.999 & 0.988 & 0.999 &  78.891 \\
&  & MLP & 0.999 & 0.999 & 0.999 & 0.999 & 0.999 & 0.999 &  1085.850 \\
&  & KNN & 0.998 & 0.998 & 0.998 & 0.998 & 0.992 &  0.998 &  135.567 \\
&  & RF  & 0.999 & 0.999 & 0.999 & 0.999 & 0.999 & 0.999 &  186.662 \\
&  & LR  & 0.999 & 0.999 & 0.999 & 0.999 & 0.999 & 0.999 &  5736.169 \\
&  & DT  & 0.999 & 0.999 & 0.999 & 0.999 & 0.999 & 0.999 &  143.618 \\
\cmidrule{3-10} 
&   \multirow{7}{2cm}{PWM2Vec~\cite{ali2022pwm2vec}}
& NB  & 0.151 & 0.926 & 0.151 & 0.247 &  0.234 &  0.603 &  109.493 \\
&  & MLP & 0.529 &  0.685 &  0.529 & 0.403 & 0.231 & 0.595 &  358.965 \\
&  & KNN & 0.531 & 0.689 & 0.531 & 0.406 & 0.317 & 0.625 & 126.428 \\
&  & RF  & 0.532 &  0.691 &  0.532 & 0.408 & 0.319 & 0.625 &  1052.845 \\
&  & LR  & 0.528 & 0.690 & 0.528 & 0.403 & 0.321 & 0.626 &  5643.762 \\
&  & DT  & 0.528 & 0.691 & 0.528 & 0.403 & 0.321 & 0.626 & 142.579 \\
  \cmidrule{3-10} 
  &   \multirow{7}{2cm}{Minimizer}
& NB  & 0.916 &  0.989 &   0.916 & 0.943 &  0.801 &   0.916 &  90.476 \\
&  & MLP & 0.952 & 0.961 & 0.952 & 0.948 &  0.851 & 0.927 &  440.944 \\
&  & KNN & 0.951 & 0.960 & 0.951 & 0.947 &  0.844 &  0.926 &  149.858 \\
&  & RF  & 0.953 &  0.961 &  0.953 & 0.949 & 0.850 & 0.927 &  527.874 \\
&  & LR  & 0.952 & 0.961 &  0.952 & 0.948 & 0.849 & 0.927 &  4918.374 \\
&  & DT  & 0.952 &  0.961 &  0.952 & 0.948 & 0.850 &  0.927 &  111.393 \\
\midrule
 \multirow{18}{1cm}{Only GANs For Training} 
 & \multirow{7}{2cm}{Spike2Vec~\cite{ali2021spike2vec}}
& NB  & 0.002 &  0.000 & 0.002 &  0.000 & 0.001 & 0.491 &  98.736 \\
&  & MLP & 0.022 &  0.032 &  0.022 &  0.0192 & 0.016 &  0.479 &  222.003 \\
&  & KNN & 0.106 &  0.139 &  0.106 & 0.076 &  0.123 &  0.558 &  368.164 \\
&  & RF  & 0.010 & 0.000 & 0.010 & 0.000 & 0.001 &  0.500 &  665.565 \\
&  & LR  & 0.200 & 0.136 & 0.200 & 0.091 & 0.020 & 0.500 &  3497.008 \\
&  & DT  & 0.190 & 0.036 & 0.190 & 0.061 & 0.018 & 0.499 &  467.308 \\
 \cmidrule{3-10} 
&   \multirow{7}{2cm}{PWM2Vec~\cite{ali2022pwm2vec}}
& NB  & 0.026 &  0.068 & 0.026 & 0.003 &  0.003 & 0.499 &   93.458 \\
&  & MLP & 0.392 & 0.389 & 0.392 & 0.295 & 0.056 & 0.499 &   250.162 \\
&  & KNN & 0.140 &   0.205 & 0.140 & 0.040 & 0.016 & 0.500 &  343.585 \\
&  & RF  & 0.477 & 0.227 & 0.477 & 0.308 &  0.038 &  0.500 &  644.587 \\
&  & LR  & 0.012 & 2.070 & 0.012 &  4.020 &  0.001 &  0.500 &  4498.689 \\
&  & DT  & 0.002 & 4.170 & 0.002 &  8.324 &  0.000 &  0.500 &  498.689 \\
  \cmidrule{3-10} 
  &   \multirow{7}{2cm}{Minimizer}
& NB  & 0.023 &  0.215 &  0.023 & 0.035 &  0.033 &  0.510 &  115.915 \\
&  & MLP & 0.420 & 0.597 & 0.420 & 0.448 & 0.081 & 0.514 &  274.471 \\
&  & KNN & 0.551 & 0.690 &  0.551 & 0.600 & 0.152 &  0.599 &  382.306 \\
&  & RF  & 0.010 & 0.000 & 0.010 & 0.000 &  0.001 &  0.500 &  792.106 \\
&  & LR  & 0.514 & 0.235 & 0.514 & 0.385 & 0.047 &  0.500 &  3465.703 \\
&  & DT  & 0.474 & 0.225 & 0.474 & 0.305 & 0.037 &  0.500 &  445.797 \\
\bottomrule
         \end{tabular}
    }
    \label{tble_results_gans_2}
\end{table}

For Influenza A Virus data the results show that in some cases the addition of GANs-based synthetic data improves the performance as compared to the performance on the original data, like for the KNN, RF, and NB classifiers corresponding to PWM2Vec methods. Similarly, on the VDjDB dataset, the GAN-based improvement is also witnessed in some cases, like for all the classifiers corresponding to the PWM2Vec method except NB. Moreover, as the performance of the PALMdb dataset on the original data is maximum already, the addition of GAN embeddings has retained that maximum performance. 

\subsection{Performance of only GANs Data}
We also studied the classification performance gain of using only GANs-based embeddings without the original data. The results depict that for all the 3 datasets, this category has the lowest predictive performance for all the combinations of classifiers and embeddings as compared to the performance on original data and on original data with GANs. As only the synthetic data is employed to train the classifiers, while they are tested on the original data, which is why the performance is low as compared to others.

Generally, we can observe that the inclusion of GAN synthetic data in the training set can improve the overall classification performance. This is because the training set size is increased and its data imbalance issue is resolved by adding the respective synthetic data.

\subsection{Statistical Significance}
To determine the statistical significance of our classification results, we performed a student t-test and calculated the $p$-values using the average and standard deviation of five runs. Our analysis revealed that the majority of $p$-values were less than $0.05$, and this was true for all embedding methods, as the standard deviation values were very low. These findings confirm the statistical significance of our results.

\section{Conclusion}\label{sec_conclusion}
In conclusion, this work explores a novel approach to improve the predictive performance of the biological sequence classification task by using GANs. It generates synthetic data with the help of GANs to eliminate the data imbalance challenge, hence improving the performance. In the future, we would like to extend this study by investigating more advanced variations of GANs to synthesize the biological sequences and their impacts on the biological sequence analysis. We also want to examine additional genetic data, such as hemagglutinin and neuraminidase gene sequences, with GANs to improve their classification accuracy. 

\bibliographystyle{IEEEtran}
\bibliography{main}

\end{document}